\documentclass[10pt,twocolumn,letterpaper]{article}

\usepackage{iccv}
\usepackage{times}
\usepackage{epsfig}
\usepackage{graphicx}
\usepackage{amsmath}
\usepackage{amssymb}

\usepackage{graphicx}
\usepackage{amsmath}
\usepackage{amssymb}
\usepackage{booktabs}
\usepackage{pifont}
\usepackage{multirow}
\usepackage{arydshln}
\usepackage[outercaption]{sidecap}    
\usepackage{color, colortbl}
\usepackage[dvipsnames]{xcolor}
\usepackage[title]{appendix}
\usepackage{enumitem}
\usepackage{bm}
\definecolor{LightGray}{rgb}{0.92,0.92,0.92}
\definecolor{Red}{rgb}{1.0, 0.13, 0.32}

\usepackage{arydshln}
\usepackage{subcaption}
\usepackage{adjustbox}
\usepackage{wrapfig}

\newcommand{\eat}[1]{}
\usepackage{amsmath,amsfonts,bm}

\def\eqref#1{equation~\ref{#1}}

\def\1{\bm{1}}

\def\rve{{\mathbf{e}}}

\DeclareMathAlphabet{\mathsfit}{\encodingdefault}{\sfdefault}{m}{sl}
\SetMathAlphabet{\mathsfit}{bold}{\encodingdefault}{\sfdefault}{bx}{n}

\usepackage[pagebackref=true,breaklinks=true,letterpaper=true,colorlinks,bookmarks=false]{hyperref}

\usepackage[capitalize]{cleveref}
\crefname{section}{Sec.}{Secs.}
\Crefname{section}{Section}{Sections}
\Crefname{table}{Table}{Tables}
\crefname{table}{Tab.}{Tabs.}

\iccvfinalcopy 

\def\iccvPaperID{6990} 

\ificcvfinal\fi

\definecolor{Red}{rgb}{1.0, 0.13, 0.32}
\definecolor{mygreen}{rgb}{0.0, 0.5, 0.0}

\begin{document}

\title{Towards Models that Can See and Read}

\author{
Roy Ganz\thanks{Work done during an Amazon internship.}\\
Technion, Israel\\
{\tt\small ganz@cs.technion.ac.il}
\and
Oren Nuriel\\
AWS AI Labs\\
{\tt\small onuriel\@amazon.com}
\and
Aviad Aberdam\\
AWS AI Labs\\
{\tt\small aaberdam@amazon.com}
\and
\and
Yair Kittenplon\\
AWS AI Labs\\
{\tt\small yairk@amazon.com}
\and
Shai Mazor\\
AWS AI Labs\\
{\tt\small smazor@amazon.com}
\and
Ron Litman\\
AWS AI Labs\\
{\tt\small litmanr@amazon.com}
}

\maketitle
\ificcvfinal\thispagestyle{empty}\fi

\newcommand{\AlgoName}{UniTNT }
\newcommand{\AlgoNameNoSpace}{UniTNT}
\newcommand{\AlgoNameBLIP}{$\text{UniTNT}_{\text{BLIP}}$ }
\newcommand{\AlgoNameNoSpaceBLIP}{$\text{UniTNT}_{\text{BLIP}}$}
\newcommand{\AlgoNameALBEF}{$\text{UniTNT}_{\text{ALBEF}}$ }
\newcommand{\AlgoNameNoSpaceALBEF}{$\text{UniTNT}_{\text{ALBEF}}$}

\newcommand{\TextVqaTask}{scene-text VQA }
\newcommand{\TextVqaTaskNoSpace}{scene-text VQA}
\newcommand{\TextCapsTask}{scene-text CAP }
\newcommand{\TextCapsTaskNoSpace}{scene-text CAP}
\newcommand{\StvqaDataset}{ST-VQA }
\newcommand{\StvqaDatasetNoSpace}{ST-VQA}

\vspace*{-0.75cm}
\begin{abstract}
\vspace*{-0.5em}
Visual Question Answering (VQA) and Image Captioning (CAP), which are among the most popular vision-language tasks, 
have analogous scene-text versions that require reasoning from the text in the image. 
Despite their obvious resemblance, the two are treated independently and, as we show, yield task-specific methods that can either see or read, but not both. 
In this work, we conduct an in-depth analysis of this phenomenon and propose \AlgoNameNoSpace, a Unified Text-Non-Text approach, which grants existing multimodal architectures scene-text understanding capabilities. 
Specifically, we treat scene-text information as an additional modality, fusing it with any pretrained encoder-decoder-based architecture via designated modules.
Thorough experiments reveal that \AlgoName leads to the first single model that successfully handles both task types.
Moreover, we show that scene-text understanding capabilities can boost vision-language models' performance on general VQA and CAP by up to $2.69\%$ and $0.6$ CIDEr, respectively.
\end{abstract}

\begin{figure}[t]
    \vspace{-0.5cm}
    \centering
    \includegraphics[width=0.99\columnwidth]{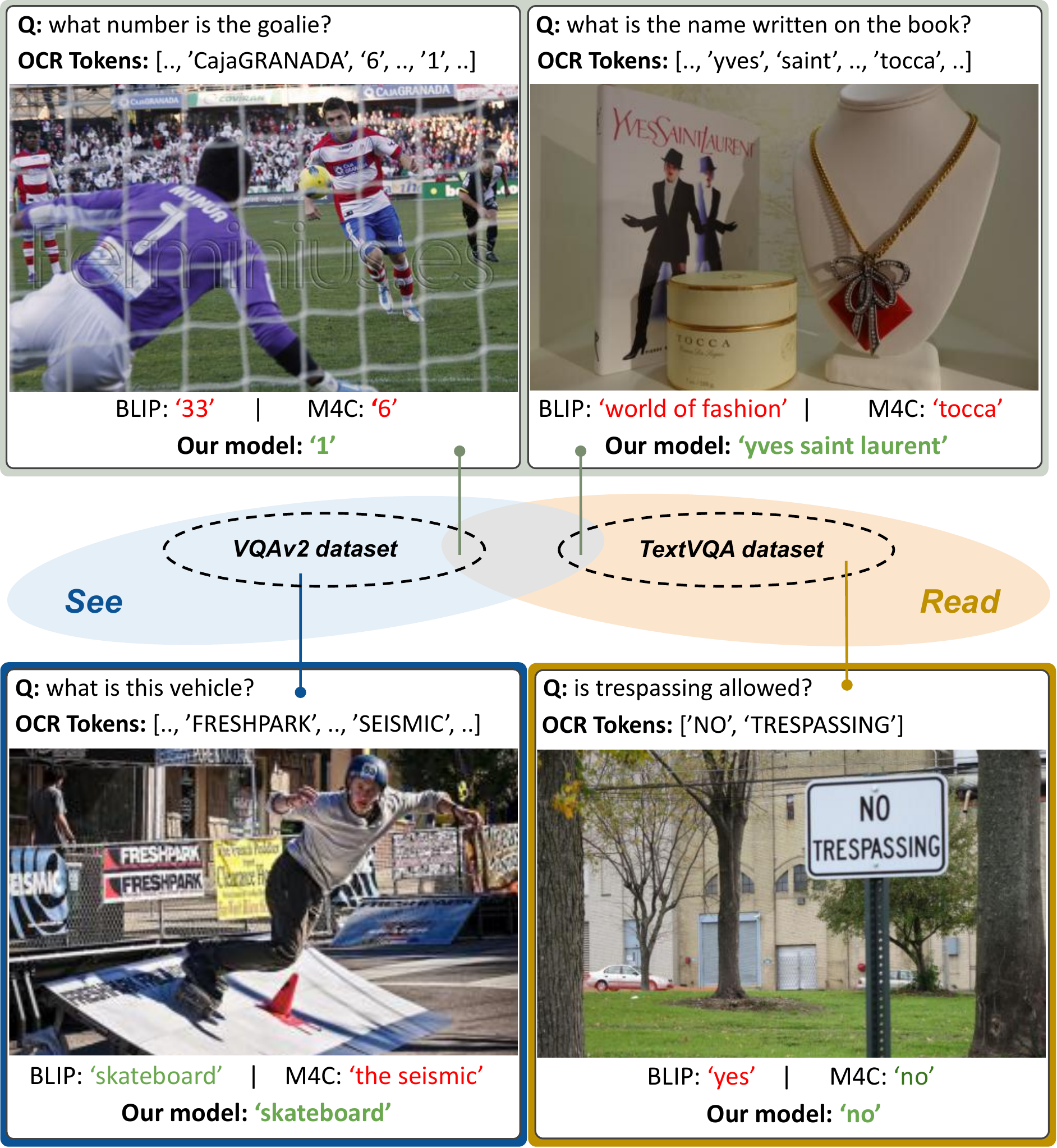}
    \caption{\textbf{See and read in VQA.} Illustration of the possible three types of reasoning required in VQA image-question pairs and representative datasets distributions (middle). Samples from the 'see' (bottom left), 'read' (bottom right), and 'see-$\boldsymbol{\cap}$-read' (top) subsets are presented. Each sample includes an image, question, OCR, and model predictions.
    }
    \label{fig:highlevel_examples_teaser}
    \vspace{-0.5cm}
\end{figure}

\vspace{-0.8cm}
\section{Introduction}
\label{sec:intro}
In recent years, Vision-Language (VL) tasks, such as Visual Question Answering (VQA)~\cite{antol2015vqa,goyal2017making} and Image Captioning (CAP)~\cite{lin2014microsoft,agrawal2019nocaps}, have gained immense research interest~\cite{zhang2021vinvl,radford2021learning,wang2021simvlm,li2022blip,hu2022scaling, wang2022git,chen2022pali, ofa}.
However, despite the remarkable success of VL models on these tasks, it was discovered a few years ago that such models are incapable of reasoning from the text in natural images~\cite{singh2019towards, biten2019scene, sidorov2020textcaps}. This finding raised significant concerns, as understanding scene-text is crucial in almost any real-world application.

To address this issue, designated scene-text datasets were introduced for both VQA \cite{singh2019towards, biten2019scene} and CAP \cite{sidorov2020textcaps}, aiming to highlight the importance of utilizing textual information in images.
Following the introduction of the above datasets, a new line of research has arisen, focusing on scene-text-oriented tasks, evaluated individually and effectively dissociated from the general one. From a user perspective, this separation is artificial and does not adequately reflect the objective of real-world VQA systems, and as we show, it encourages models to only excel on one task at a time. Therefore, we advocate that VL research should strive towards unified models, and thus, methods should be evaluated accordingly.
To this end, we propose conducting \emph{combined evaluation} for VL models on both general and scene-text benchmarks and treating the average results as expressing the ``see'' and ``read'' capabilities.
We emphasize that even the minority of works that evaluate both types of tasks~\cite{wang2022git,pali,alayrac2022flamingo} do it on separate models, which are finetuned per task, perpetuating the faulty tasks' segregation.

Apart from being unjustified, this separation introduces biases~\cite{biten2022latr,wang2021towards}, providing the models with prior knowledge that implies which modality to focus on, which does not exist in real-world scenarios. Namely, it creates a shortcut that encourages models to excel solely on a specific benchmark by acquiring an understanding of either the visual or textual information in the image, but not both.
In particular, Biten \etal~\cite{biten2022latr} recently showed that SOTA performance on \TextVqaTask can be achieved without using the visual modality, and Wang \etal~\cite{wang2021towards} revealed that existing \TextVqaTask models' success stems from exploiting language priors.
Our \emph{combined evaluation} effectively addresses this problem by testing whether models can reason from both types of information, as exploiting such data biases and priors would yield low combined results.

\begin{figure}[t]
    \centering
    \includegraphics[width=0.99\columnwidth]{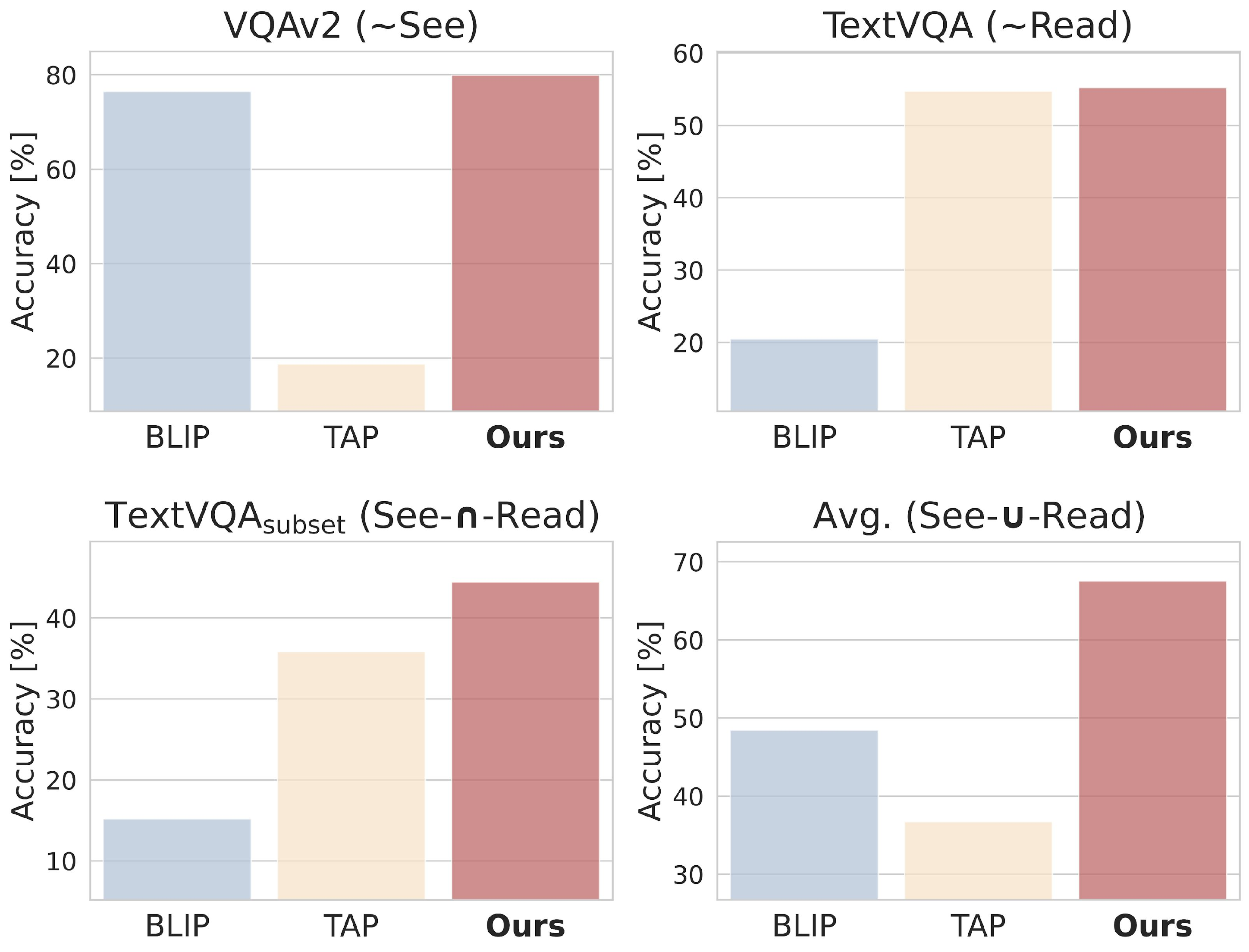}
    \caption{\textbf{Models’ accuracy on different types of VQA data.} 
    Leading methods and \AlgoName performance on different benchmarks.
    VQAv2 and TextVQA datasets mostly require reasoning from visual information only (‘See’) and textual information only (‘Read’), respectively.
    ‘See-$\boldsymbol{\cap}$-Read' refers to a subset of the TextVQA dataset (\cref{sec:subset_section}), in which both modalities are essential for answering each question. ‘See-$\boldsymbol{\cup}$-Read’ represents the sets' union.
    }
    \label{fig:barplot_secondary_teaser}
    \vspace{-0.5cm}
\end{figure}

From a more high-level view, three categories span the space of VL data; the first are examples that require reasoning over vision only (dominant in VQA~\cite{goyal2017making} and CAP~\cite{coco_cap}), the second are instances in which using scene-text information solely is sufficient (dominant in \TextVqaTaskNoSpace~\cite{biten2019scene,singh2019towards} and \TextCapsTaskNoSpace~\cite{sidorov2020textcaps}), and the third are ones in which both are essential.
We denote the three subsets as \textbf{'see'}, \textbf{'read'}, and \textbf{'see-$\boldsymbol{\cap}$-read'}, respectively. For completion, the whole space is denoted as \textbf{'see-$\boldsymbol{\cup}$-read'}, the union of all others.
We illustrate this conceptual data distribution for VQA in \cref{fig:highlevel_examples_teaser}.
Examining the performance of existing VQA approaches over the three types of questions mentioned above, shown in \cref{fig:barplot_secondary_teaser}, reveals that while some of the methods~\cite{li2022blip, albef, ofa} perform well on the first subset and some~\cite{hu2020iterative,yang2021tap} on the second, none are optimal on the entire domain. Moreover, throughout our analysis, we reveal that the 'see-$\boldsymbol{\cap}$-read' subset, in which both visual and textual information are needed for answering, is very challenging and underrepresented, requiring a new dedicated benchmark.

In this work, while striving towards models that excel on the entire space of VL data, we propose \AlgoNameNoSpace, a Unified Text-Non-Text model, which provides VL architectures with scene-text understanding capabilities.
Specifically, we treat textual information in the image, \textit{i.e.} tokens and positions, as a third modality and introduce it into the pretrained model.
Adding a new modality to an already-trained model is challenging and might lead to suboptimal results~\cite{french1999catastrophic,tsimpoukelli2021multimodal,zhai2022lit,alayrac2022flamingo}. 
To overcome this, we encode such information using a designated encoder and inject it into the existing pretrained decoder via a novel fusing mechanism that gradually shifts between VL features to textual-enriched ones.
Moreover, we propose scene-text-related intermediate supervision to encourage the already-trained model to leverage the newly added information. 
Being both task and model agnostic by its design, our method can be applied to any VL encoder-decoder-based architecture.

We evaluate \AlgoName on both general and scene-text benchmarks of VQA and CAP using \emph{combined evaluation} and show that it leads to the first single model performing well on both tasks.
We show that our method can be easily integrated into existing VL models, improving their scene-text understanding substantially by applying it to BLIP~\cite{li2022blip} and ALBEF~\cite{albef}. 
Interestingly, such reasoning abilities boost the base model's VQA results (\textit{e.g.}, improves BLIP~\cite{li2022blip} by $2.69\%$ on VQAv2~\cite{goyal2017making}), while achieving state-of-the-art competitive results on \TextVqaTask benchmarks.
A similar trend exists in captioning, where \AlgoName enhances BLIP's performance by 0.6 CIDEr points on COCO Captions~\cite{coco_cap} while substantially boosting its \TextCapsTask performance.
These improvements highlight the significance of scene-text comprehension in VL tasks, laying the foundation for future research on general multimodal architectures that can leverage scene-text.

\begin{table*}[!t]
\centering
\begin{adjustbox}{max width=\textwidth}
\begin{tabular}{l l c| c c c c c | c || c c c | c}
     \toprule
     & \multirow{3}{*}{\textbf{Method}} & \multirow{3}{*}{\shortstack[c]{\textbf{OCR} \\ \textbf{System}}} &  \multicolumn{6}{c||}{\textbf{Visual Question Answering}} & \multicolumn{4}{c}{\textbf{Image Captioning}} \\
     & & & \multicolumn{2}{c}{\textbf{VQA}} & \multicolumn{2}{c}{\textbf{TextVQA}} & \textbf{ST-VQA} & \multirow{2}{*}{\textbf{Avg.}} & \textbf{COCO} & \multicolumn{2}{c|}{\textbf{TextCaps}} & \multirow{2}{*}{\textbf{Avg.}} \\
     & & & test-dev & test-std & val & test & test-ANLS & & Karpathy-test & val & test & \\
     \hline
     \parbox[t]{2mm}{\multirow{5}{*}{\rotatebox[origin=c]{90}{\textbf{Separate}}}}
     & M4C~\cite{hu2020iterative,sidorov2020textcaps} & \checkmark & 27.47 & 27.70 & 46.53 & 47.42 & 0.43 & 37.56 & 4.7 & 95.5 & 90.1 & 47.4 \\
     & TAP~\cite{yang2021tap} & \checkmark & 18.76 & 18.81 & 54.71 & 53.97 & 0.60 & 36.39 & 4.6 & 109.2 & 103.2 & 53.9  \\
     & ALBEF~\cite{albef} & \ding{55} & 75.22 & 75.38 &  11.67 & 13.88 & 0.19 & 44.63 & - & - & - & -\\
     & BLIP~\cite{li2022blip} & \ding{55} & 76.39 & 76.59 &  20.50 & 23.74 & 0.34 & 50.16 & 133.3 & 59.4 & 61.9 & 97.6 \\
     & $\text{OFA}_{\text{Large}}$~\cite{ofa} & \ding{55} & 79.70 & 79.85 &  22.10 & 21.47 & 0.27 & 50.66 & 150.7 & 64.5 & 66.8 & 108.8  \\
     \hline
     \parbox[t]{2mm}{\multirow{3}{*}{\rotatebox[origin=c]{90}{\textbf{ Comb.}}}}
     & M4C \cite{hu2020iterative} & \checkmark & 59.11 & 59.04 & 47.22 & 48.61 & 0.50 & 53.83 & 109.8 & 102.7 & 98.0 & 103.9  \\
     & ALBEF \cite{albef} & \ding{55} & 75.61 & 75.87 & 16.15 & 17.04 & 0.22 & 46.46 & - & - & - & -\\
     & BLIP \cite{li2022blip} & \ding{55} & 77.40 & 77.39 &  32.43 & 31.48 & 0.44 & 54.44 & 133.4 & 101.4 & 91.8 & 112.6 \\
\bottomrule
\end{tabular}
\end{adjustbox}
\caption{\textbf{Current status of VQA and CAP models.}
The results of leading methods on both scene-text and general VQA and CAP benchmarks reveal that currently, no method performs well on both scene-text and general benchmarks, even when applying combined training.
\textit{Separate} and \textit{Comb.} summarize the results described in Sections \ref{sec:vqa_study}, \ref{sec:cap_study} and \ref{sec:comb_study}, respectively.
}
\label{tab:current_status}
\vspace{-0.5cm}
\end{table*}

To summarize:
\begin{itemize}[nolistsep,leftmargin=*]
    \item We thoroughly analyze current methods and reveal that the faulty text-non-text task separation leads to models that either reason from visual or textual information in images, but not both.
    \item We introduce \AlgoNameNoSpace, a model-agnostic method to grant reading capabilities to pretrained VL models by fusing the scene-text information as an additional modality.
    \item Extensive experiments show that our method not only improves the scene-text benchmarks' results but also significantly enhances the performance of VQA and CAP.
\end{itemize}

\section{See and Read: Analyzing Methods and Data}
\label{sec:current_status}
In this paper, contrary to the common practice in VL research, we highlight the importance of models to ``see'' and ``read'' altogether and start by comprehensively analyzing such capability via a "see-$\boldsymbol{\cup}$-read"-oriented \emph{combined evaluation}. 
Our analysis reveals that existing models' reasoning abilities over both types of information are lacking, prompting the question of whether this limitation is due to inherent method constraints or biased data.
Our evaluation focuses on the performance of leading general and scene-text-oriented models on VQAv2~\cite{goyal2017making}, TextVQA~\cite{singh2019towards}, and \StvqaDatasetNoSpace~\cite{biten2019scene} for VQA, and COCO Captions~\cite{coco_cap} and TextCaps~\cite{sidorov2020textcaps} for captioning.


\label{sec:curr_status}
\subsection{Visual Question Answering}
\label{sec:vqa_study}
\noindent\textbf{General VQA Methods}:
During the vision-language revolution, numerous methods~\cite{li2020unimo,radford2021learning,albef,li2022blip,ofa,wang2021simvlm,wang2022git,alayrac2022flamingo,chen2022pali,zhang2021vinvl,yu2022coca,chowdhery2022palm,hu2022scaling} have been proposed for various multimodal tasks, including VQA, which have advanced the state-of-the-art. These methods can leverage vast online image-caption pairs via vision-language pretraining~\cite{align_before_fuse,uniter,radford2021learning}, followed by task-specific fine-tuning. However, a few years ago, such models were shown to be ineffective in reasoning from textual information in the scene, as they primarily focus on the images' visual content~\cite{singh2019towards, biten2019scene}. 

Nevertheless, such models have advanced significantly in the past few years. Thus, to reveal the current status of such models in scene-text understanding, we examine the performance of three leading VQA models, ALBEF~\cite{albef}, BLIP~\cite{li2022blip}, and OFA~\cite{ofa}, using unconstrained open-vocabulary generation, on scene-text VQA tasks.
As seen in \cref{tab:current_status}, although such methods perform well on VQA, as expected, their results on the analogous \TextVqaTask datasets are unsatisfactory, testifying their incompetence in scene-text understanding.
Interestingly, their inability to utilize scene-text information hinders its performance even on VQA, as we later show in \cref{sec:experiments}.

\noindent\textbf{Scene-Text VQA Methods}:
Several methods have been proposed to improve the scene-text understanding of VQA models~\cite{hu2020iterative,han2020finding,gao2021structured,yang2021tap,kant2020spatially,lu2021localize,biten2022latr}. These models utilize an off-the-shelf OCR system's output alongside the image and question as input to a multimodal transformer. However, some recent studies~\cite{wang2021towards,biten2022latr} have indicated that \TextVqaTask datasets may have biases discouraging models from relying on the visual modality. 
To properly test such claims, we evaluate M4C~\cite{hu2020iterative} and TAP~\cite{yang2021tap} on general VQA, which requires strong visual understanding and report the results in~\cref{tab:current_status}.
As can be seen, M4C and TAP obtain only $27.70\%$ and $18.81\%$, respectively. 
When compared to, for example, BLIP's $76.59\%$, it testifies that, indeed, such methods disregard the visual information.
Interestingly, although TAP consistently outperforms M4C on the scene-text benchmarks, it achieves lower results on the general one, implying the data biases in the former datasets.

\subsection{Image Captioning}
\label{sec:cap_study}
Similar to VQA's analysis, we conduct a captioning \emph{combined evaluation} using TextCaps and COCO Captions for both types of models and report the average CIDEr scores.
Our empirical results in \cref{tab:current_status} demonstrate that while general models (BLIP and OFA) and scene-text ones (M4C-Captioner and TAP) perform well on their designated benchmarks, they fail to obtain satisfactory results on the analogous one.
In particular, BLIP obtains a CIDEr score of only $61.9$ on TextCaps, compared to $90.1$ of M4C-Captioner.
On the other hand, the latter achieves $4.7$ on COCO captions, compared to BLIP's $133.3$.
In addition, like in VQA, while TAP outperforms M4C in TextCaps, it does not occur on COCO Captions.
These findings suggest that existing methods exhibit unsatisfactory performance when evaluated on both captioning benchmarks.

\begin{figure*}[htp!]
    \centering
    \includegraphics[width=0.99\textwidth]{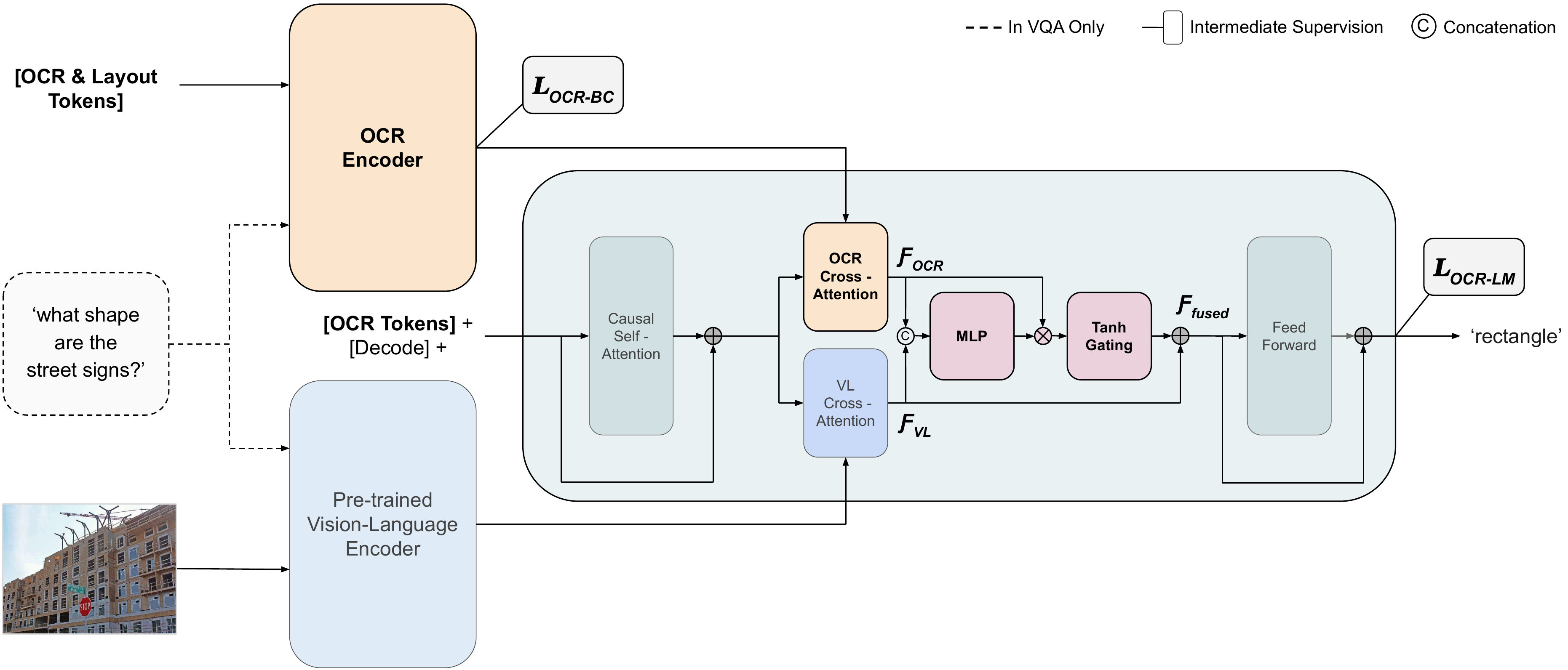}
    \caption{\textbf{An overview of \AlgoNameNoSpace.} Our method endows existing general VL models with scene-text understanding capability.
    The OCR information is encoded separately and injected into the decoder via a gated cross-attention-based fusing mechanism as complemental information. $\mathcal{L}_{\text{OCR-BC}}$ and $\mathcal{L}_{\text{OCR-LM}}$ are auxiliary losses, enforcing the model to utilize the scene-text information. \AlgoName newly introduced components are presented in bold. ‘See’, ‘Read’, and 'Fusing' related modules are in blue, orange, and red, respectively.}
    \label{fig:main_figure}
    \vspace{-0.5cm}
\end{figure*}

\subsection{The Role of the Datasets in 'See and Read'}
\label{sec:comb_study}
We now examine whether this limitation stems from a lack of representative training data rather than method limitations. Specifically, we test if the inferior performance of scene-text-oriented models on visual tasks and vice versa is solely due to the training data's bias towards reasoning over solely one type of information. 
To test this claim, we merge two datasets, conduct combined training for both general and scene-text-oriented methods, and report the results in \cref{tab:current_status}.
As can be seen, while unified training leads to improved performance on both types of VL benchmarks, there is a substantial performance gap -- scene-text models lag behind general ones on the general benchmarks and vice-versa. Nevertheless, these results indicate that reasoning from text and vision are not at odds and suggest a symbiotic relationship between the two tasks. Furthermore, they provide further motivation for avoiding the common practice of separating the tasks, as done in previous work~\cite{alayrac2022flamingo,pali,wang2022git}.
To conclude, while joint training is a step forward, it is not enough to achieve our ultimate goal.


\section{Method}
In this section, we describe \AlgoNameNoSpace, a method aimed to obtain our titular goal by granting pretrained general VL models the ability to reason over scene-text information during finetuning while retaining their original reasoning capabilities, depicted in \cref{fig:main_figure}. 
By doing so, we propose a change of perspective compared to top-performing ST methods, such as~\cite{biten2022latr,yang2021tap}, that harnesses an OCR-oriented pretrained model but fails to enrich it with visual understanding during finetuning.
Adapting pretrained models to consider additional inputs, absent during pretraining, is a non-trivial task tackled by recent literature~\cite{tsimpoukelli2021multimodal, alayrac2022flamingo}. On the one hand, we wish to encourage the model to utilize the new stream of information and, on the other hand, to prevent it from neglecting the original stream. To address this, we encode the OCR information via a designated OCR encoder and fuse it residually, retaining the former stream of information and gradually shifting towards an OCR-enriched one. Moreover, we propose auxiliary losses, encouraging the pretrained decoder to utilize this information. 
Similarly to previous works~\cite{hu2020iterative,yang2021tap,lu2021localize,biten2022latr}, we utilize an off-the-shelf-OCR system to extract the scene-text information. 

\subsection{Architecture}
\label{sec:arch}
We design our architecture in a task-agnostic way -- enabling compatibility with both visual question answering and image captioning tasks. In addition, \AlgoName is model agnostic and can be applied to any encoder-decoder-based VL model. In this work, we integrate our approach into two top-performing open-source methods -- ALBEF \cite{albef}, and BLIP \cite{li2022blip} as a case study, denoted as $\text{UniTNT}_{\text{ALBEF / BLIP}}$.


\paragraph{OCR Encoder}
Rather than utilizing the pre-existing encoder to process the OCR alongside the visual modality, as in~\cite{biten2022latr,hu2020iterative,yang2021tap}, we introduce a dedicated OCR encoder, which maps the scene-text information into features fed into the existing system's decoder. This encoder receives the question alongside OCR information, namely tokens and 2-dimensional (2D) positional information, both extracted by the OCR system. The positional information was proven to be valuable for documents and scene-text understanding tasks~\cite{xu2020layoutlm,xu2020layoutlmv2,appalaraju2021docformer,biten2022latr}.
Not only that our approach outperforms the one that utilizes the pre-existing text encoder to process the OCR tokens (demonstrated in \cref{sec:ablations}), but it also provides flexibility to address tasks that do not utilize a text encoder, such as image captioning.

Formally, each OCR instance is represented by $(t,x_0,y_0,x_1,y_1,w,h)$, namely, its word token, bounding box's top-left, bottom-right, width, and height values, respectively.
We embed each value separately using designated embedding layers $\operatorname{E}$ (\textit{i.e.}, \texttt{torch.nn.Embedding}).
Next, we sum the 2D representations, pass them via a 2-layer MLP and add it to the token's representation, yielding the OCR representation:
\begin{equation}
\label{eq:2d}
\begin{split}
\rve_{\text{OCR}} = & \operatorname{E_{OCR}(t)} + \alpha * \operatorname{MLP}(\operatorname{E_x}(x_0) + \operatorname{E_y}(y_0) + \\ 
                    & \operatorname{E_x}(x_1) + \operatorname{E_y}(y_1) + \operatorname{E_w}(w) + \operatorname{E_h}(h))
\end{split}
\end{equation}
where $\alpha$ is a predefined hyperparameter.
As for the question, we embed its tokens using the same embedding layer.
Since both the OCR and the question representations are fed into the same model, we equip the question representations with pseudo-2D information corresponding to the size of the entire image, yielding the final question representation $\rve_{q}$.
Finally, we concatenate them to obtain the OCR encoder's input, $\left\{\rve_{q}^1 \dots \rve_{q}^M, \rve_\text{OCR}^1 \dots \rve_\text{OCR}^N\right\}$, where M and N are the lengths of the question and OCR, respectively.

\paragraph{VL-OCR Decoder}
To integrate the OCR information into the decoder, we add a dedicated OCR Cross Attention (CA) and a fusing mechanism, as visualized in \cref{fig:main_figure}.
We place the OCR CA block parallel to the pre-existing VL CA module to enrich the decoded features with textual information in the image.
This architectural design yields two data streams (visual and scene-text-oriented ones) that need to be merged adequately into a single VL-OCR representation.
To this end, we introduce a fusing mechanism composed of a gated cross-attention mechanism, which gradually shifts from VL features to fused, OCR-enriched ones.

Formally, our fusing mechanism merges the output of our new OCR CA with the one of the VL CA, denoted as $\mathcal{F}_{\text{OCR}}$ and $\mathcal{F}_{\text{VL}}$ respectively.
Specifically, this module receives two features sequences, ${\mathcal{F}_{\text{OCR}}, \mathcal{F}_{\text{VL}} \in \mathbb{R}^{B\times L\times C}}$, and outputs ${\mathcal{F}_{\text{fused}} \in \mathbb{R}^{B\times L\times C}}$, where $B, L, C$ are the batch size, sequence length and the number of channels, respectively.
First, we concatenate $\mathcal{F}_{\text{OCR}}$ and $\mathcal{F}_{\text{VL}}$ across the channel dimension and insert them into a simple 2-layer MLP to obtain an attention map $\mathcal{F}_\text{attn}\in \mathbb{R}^{B\times L\times C}$.
Next, we pass the element-wise product of $\mathcal{F}_{\text{OCR}}$ and $\mathcal{F}_\text{attn}$ in a $\tanh$ gating mechanism~\cite{lstm,alayrac2022flamingo}.
The goal of the $\tanh$ gating is to enable gradual OCR blending with the VL one by multiplying its inputs with $\tanh(\beta)$, where $\beta$ is a learnable parameter initialized to zero.
At initialization, it ensures that the added modules are skipped, preserving the model pretraining's data flow.
Finally, we sum the output of the $\tanh$ gating with $\mathcal{F}_{\text{VL}}$ to obtain the fused features:
\begin{equation}
    \mathcal{F}_\text{attn} = \operatorname{MLP}(\operatorname{concat}(\mathcal{F}_{\text{VL}}, \mathcal{F}_{\text{OCR}})),
\end{equation}
\begin{equation}
    \mathcal{F}_{\text{fused}} = \mathcal{F}_{\text{VL}} + \tanh(\beta) (\mathcal{F}_{\text{OCR}}\odot \mathcal{F}_\text{attn}),
\end{equation}
where $\odot$ is the Hadamard product.

\subsection{Scene-text Auxiliary Losses}
We propose two auxiliary losses, encouraging the model to utilize the scene-text signal rather than ignoring it - OCR Causal Language Modeling (OCR-LM) and OCR Binary Classification (OCR-BC).

\paragraph{OCR Causal Language Modeling}
To better fuse the scene-text information, we add a causal language modeling supervision over the OCR tokens. Specifically, we prepend the shifted OCR tokens (according to the OCR system reading order) to the inputs of the decoder and train the system to predict the next OCR token based on previous ones,
\begin{equation}
    \mathcal{L}_{\text{OCR-LM}} = -\sum_{i=1}^{N} \operatorname{log}\left(\mathbb{P}\left(t^i|t^{<i}\right)\right)
\end{equation}
where $t^i$ is the $i$\textsuperscript{th} OCR token.
Minimizing such loss enforces the system to account for the scene-text signal, as desired.
While variants of such a loss were previously used during pretraining~\cite{yang2021tap, biten2022latr}, we are the first to utilize it during finetuning.
Moreover, inserting the OCR into the decoder at inference has another significant advantage, as it serves as a prefix and enables the model to condition its answers on the OCR.
Such behavior is desirable since the OCR can provide meaningful information for general and scene-text VL tasks, as we experimentally demonstrate in \cref{sec:vqa_exp}.

\paragraph{OCR Binary Classification}
To obtain more meaningful and task-beneficial OCR encodings, we propose a binary classification objective of predicting whether each OCR token is a part of the ground-truth answer.
We build a binary linear classifier on top of the outputs of the OCR encoder and train it using a binary cross-entropy loss.
More specifically, since most of the OCR tokens are not part of the answer, we employ a weighted version, as such classification task is highly imbalanced.
We denote this loss as $\mathcal{L}_{\text{OCR-BC}}$.

\begin{table}[!t]
	\centering
	\resizebox{1\linewidth}{!}{%
	\begin{tabular}	{@{\extracolsep{1pt}}l l c | c c c c c | c}
        \toprule	 	
        & \multirow{2}{*}{\textbf{Method}} & \multirow{2}{*}{\shortstack[c]{\textbf{OCR} \\ \textbf{System}}} &  \multicolumn{2}{c}{\textbf{VQA}} & \multicolumn{2}{c}{\textbf{TextVQA}} & \textbf{ST-VQA} & \textbf{Avg.} \\
        & & & test-dev & test-std & val & test & test-ANLS & \\
        \hline
        \parbox[t]{2mm}{\multirow{8}{*}{\rotatebox[origin=c]{90}{\textbf{\small VQA}}}} 
        & $\text{SimVLM}_{\text{large}}$ \cite{wang2021simvlm} & \ding{55} & \color{gray}{79.32} & \color{gray}{79.56} & - & - & - & - \\
        & $\text{GIT}_{\text{large}}^{\text{VQA}}$~\cite{wang2022git} & \ding{55} & 75.51 & - & - & - & - & - \\
        & ALBEF \cite{albef} & \ding{55} & 75.22 & 75.38 & 11.67 & 13.88 & 0.19 & 44.63 \\
        & $\text{OFA}_{\text{Large}}$~\cite{ofa} & \ding{55} & 79.70 & \underline{79.85} & 22.10 & 21.47 & 0.27 & 50.66 \\
        & $\text{mPLUG}_{\text{ViT-B}}$~\cite{li2022mplug} & \ding{55} & \color{gray}{\underline{79.79}} & \color{gray}{79.81} & - & - & - & - \\
        \cdashline{2-9}
        & BLIP \cite{li2022blip} & \ding{55} & 76.39 & 76.59 & 20.50 & 23.74 & 0.34 & 50.17 \\
        & \AlgoNameBLIP  & \checkmark & 79.68 & 79.78 & 36.33 & 35.90 & 0.50 & 57.84 \\
        & $\quad \Delta$  & & \color{mygreen}$\uparrow3.28$ & \color{mygreen}$\uparrow3.19$ & \color{mygreen}$\uparrow15.83$ & \color{mygreen}$\uparrow12.16$ & \color{mygreen}$\uparrow0.16$ & \color{mygreen}$\uparrow7.67$ \\
        \hline
        \parbox[t]{2mm}{\multirow{9}{*}{\rotatebox[origin=c]{90}{\textbf{\small TextVQA}}}} & $\text{GIT}_{\text{large}}^{\text{TextVQA}}$~\cite{wang2022git} & \ding{55} & - & - & 37.47 & - & - & - \\
        & SA-M4C\cite{kant2020spatially} & \checkmark & - & - & \color{gray}{45.4} & \color{gray}{44.6} & \color{gray}{0.50} & - \\
        & LOGOS \cite{lu2021localize} & \checkmark & - & - & \color{gray}{51.53} & \color{gray}{51.08} & \color{gray}{0.58} & - \\
        & M4C \cite{hu2020iterative} & \checkmark & \color{gray}{27.47} & \color{gray}{27.70} & \color{gray}{46.53} & \color{gray}{47.42} & \color{gray}{0.43} & 37.56 \\
        & TAP \cite{yang2021tap} & \checkmark & \color{gray}{18.76} & \color{gray}{18.81} & \color{gray}{54.71} & \color{gray}{53.97} & \color{gray}{0.60} & 36.39 \\
        & LaTr \cite{biten2022latr} & \checkmark & - & - & \textbf{59.53} & \textbf{59.55} & \textbf{0.68} & - \\
        \cdashline{2-9}
        & BLIP \cite{li2022blip} & \ding{55} & 40.16 & 40.39 & 30.12 & 27.72 & 0.36 & 34.06 \\
        & \AlgoNameBLIP  & \checkmark & 37.01 & 37.24 & 50.19 & 47.39 & 0.59 & 42.32 \\
        & $\quad \Delta$  & & \color{Red}$\downarrow3.15$ & \color{Red}$\downarrow3.15$ & \color{mygreen}$\uparrow20.07$ & \color{mygreen}$\uparrow19.67$ & \color{mygreen}$\uparrow0.23$ & \color{mygreen}$\uparrow8.26$ \\
        \hline
        \parbox[t]{2mm}{\multirow{7}{*}{\rotatebox[origin=c]{90}{\textbf{\small Comb.}}}}
        & M4C \cite{hu2020iterative} & \checkmark & \color{gray}{59.11} & \color{gray}{59.04} & \color{gray}{47.22} & \color{gray}{48.61} & \color{gray}{0.50} & 53.83 \\
        \cdashline{2-9}
        & ALBEF \cite{albef} & \ding{55} & 75.61 & 75.87 & 16.15 & 17.04 & 0.22 & 46.46 \\
        & $\text{\AlgoNameNoSpaceALBEF}$ & \checkmark & 77.60 & 77.80 & 43.73 & 44.13& 0.58 & \underline{60.97} \\
        & $\quad \Delta$  & & \color{mygreen}$\uparrow1.99$ & \color{mygreen}$\uparrow1.93$ & \color{mygreen}$\uparrow27.58$ & \color{mygreen}$\uparrow27.09$ & \color{mygreen}$\uparrow0.36$ & \color{mygreen}$\uparrow14.51$ \\
        & BLIP \cite{li2022blip} & \ding{55} & 77.40 & 77.39 &  32.43 & 31.48 & 0.44 & 54.44 \\
        & \AlgoNameBLIP  & \checkmark & \textbf{79.90} & \textbf{80.08} & \underline{55.21} & \underline{55.35} & \underline{0.66} & \textbf{67.72} \\
        & $\quad \Delta$  & &\color{mygreen}$\uparrow2.50$ & \color{mygreen}$\uparrow2.69$ & \color{mygreen}$\uparrow22.77$ & \color{mygreen}$\uparrow23.87$ & \color{mygreen}$\uparrow0.22$ & \color{mygreen}$\uparrow13.28$ \\
	    \bottomrule
	\end{tabular}
	}
	\caption{\textbf{VQA results.}
    Accuracy of general, scene-text oriented VQA methods and \AlgoName using three training regimes -- separate VQA and TextVQA and combined training, where non-open vocabulary methods results are in {\color{gray}gray}.
    $\Delta$ indicates improvement over the base architecture in the same regime.
    These results highlight our method's effectiveness, significantly improving the general VQA results by enriching VL models with scene-text understanding.
    }
    \label{table:vqa_textvqa}
\end{table}
\subsection{Training Procedure}
So far, we have described the main building blocks in our method, and now, as illustrated in \cref{fig:main_figure}, we put it all together.
First, we harness a trained general encoder-decoder VL model and modify it as described above in \cref{sec:arch}.
Next, we freeze the VL model's pre-existing image encoder, similarly to \cite{zhai2022lit,alayrac2022flamingo}, and train \AlgoName on a unified dataset (\textit{i.e.}, general and scene-text VQA datasets or general and scene-text captioning datasets).
Specifically, $\mathcal{L}_{\text{\AlgoName}} = \mathcal{L}_{\text{base}} + \alpha_1\mathcal{L}_{\text{OCR-LM}} + \alpha_2\mathcal{L}_{\text{OCR-BC}}$ is minimized, where $\mathcal{L}_{\text{base}}$ is the base task-dependent loss term used in our base architecture, and $\alpha_1$, $\alpha_2$ are tunable hyperparameters.

\section{Experiments}
\label{sec:experiments}
\begin{table}[!t]
	\centering
	\resizebox{1\linewidth}{!}{%
	\begin{tabular}	{@{\extracolsep{1pt}}l l c | c c c | c }
        \toprule	 	
        & \multirow{2}{*}{\textbf{Method}} & \multirow{2}{*}{\shortstack[c]{\textbf{OCR} \\ \textbf{System}}} & \textbf{COCO} &  \multicolumn{2}{c|}{\textbf{TextCaps}} & \textbf{Avg.} \\ 
         & & & Karpathy-test & val & test & \\
        \hline
        \parbox[t]{2mm}{\multirow{8}{*}{\rotatebox[origin=c]{90}{\textbf{\footnotesize Caps}}}}
        & VinVL \cite{zhang2021vinvl} & \ding{55} & 129.3 & - & - & - \\
        & $\text{LEMON}_{\text{base}}$ \cite{hu2022scaling} & \ding{55} & 133.3 & - & - & - \\
        & $\text{GIT}_{\text{large}}^{\text{Cap}}$~\cite{wang2022git} & \ding{55} & 138.5 & - & - & - \\
        & $\text{SimVLM}_{\text{large}}$~\cite{wang2021simvlm} & \ding{55} & \underline{142.6} & - & - & - \\
        & $\text{OFA}_{\text{Large}}$ \cite{ofa} & \ding{55} & \textbf{150.7} & 64.5 & 66.8 & 108.8 \\
        \cdashline{2-7}
        & BLIP \cite{li2022blip} & \ding{55} & 133.3 & 59.4 & 61.9 & 97.6 \\
        & \AlgoNameBLIP & \checkmark & 133.7 & 59.6 & 62.8 & 98.3  \\
        & $\quad \Delta$ & & \color{mygreen}$\uparrow0.4$  & \color{mygreen}$\uparrow0.2$  & \color{mygreen}$\uparrow0.9$ & \color{mygreen}$\uparrow0.7$  \\
        \hline
        \parbox[t]{2mm}{\multirow{7}{*}{\rotatebox[origin=c]{90}{\textbf{\footnotesize TextCaps}}}} & $\text{GIT}_{\text{large}}^{\text{TextCap}}$~\cite{wang2022git} & \ding{55} & - & 106.3 & - & - \\
        & MMA-SR~\cite{wang2020multimodal} &\checkmark & - & 98.0 & 88.0 & - \\
        & CNMT~\cite{anonymous2020challenge} & \checkmark & - & - & 93.0 & - \\
        & M4C-Captioner \cite{sidorov2020textcaps} & \checkmark & 4.7 & 95.5 & 90.1 & 47.4  \\
        & TAP \cite{yang2021tap} & \checkmark & 4.6 & 109.2 & 103.2 & 53.9 \\
        \cdashline{2-7}
        & BLIP \cite{li2022blip} & \ding{55} & 84.8 & 112.7 & 103.7 & 94.3 \\
        & \AlgoNameBLIP & \checkmark & 70.4 & \textbf{130.5} & \textbf{123.1} & 96.8  \\
        & $\quad \Delta$ & & \color{Red}$\downarrow14.4$  & \color{mygreen}$\uparrow17.8$  & \color{mygreen}$\uparrow19.4$ & \color{mygreen}$\uparrow2.5$  \\
        \hline
        \parbox[t]{2mm}{\multirow{4}{*}{\rotatebox[origin=c]{90}{\textbf{\footnotesize Comb.}}}}
        & M4C-Captioner \cite{sidorov2020textcaps} & \checkmark & 109.8 & 102.7 & 98.0 & 103.9  \\
        \cdashline{2-7}
        & BLIP \cite{li2022blip} & \ding{55} & 133.4 & 101.4 & 91.8 & \underline{112.6} \\
        & \AlgoNameBLIP & \checkmark & 134.0 & \underline{119.1} & \underline{109.4} & \textbf{121.7}  \\
        & $\quad \Delta$ & & \color{mygreen}$\uparrow0.6$  & \color{mygreen}$\uparrow17.7$  & \color{mygreen}$\uparrow17.6$ & \color{mygreen}$\uparrow9.1$  \\
	    \bottomrule
	\end{tabular}
	}
	\caption{\textbf{CAP results.} 
        CIDEr scores of general, scene-text oriented CAP methods and \AlgoName using three training regimes -- separate Caps and TextCaps and combined training.
        $\Delta$ indicates improvement over the base architecture in the same regime.
        These results highlight our method's effectiveness, significantly improving the general CAP results by enriching VL models with scene-text understanding.
        }
	\label{table:textcap_cap}
\end{table}

In this section, we experimentally examine \AlgoNameNoSpace, comparing its performance with state-of-the-art methods with a similar capacity on both VQA and CAP tasks, using separate and combined training. 
In particular, to better study the effects of our method, we test it and the baselines in three distinct training regimes; (i) separate training on the general datasets, (ii) separate training on the scene-text ones, and (iii) combined training approach, denoted as Comb.
As we focus on models' see and read capabilities, we emphasize the combined training regime and view it as the most crucial one.
However, the separate training regimes can provide insights into the impact of scene-text understanding on the general benchmarks and the biases within the scene-text datasets.
As in \cref{sec:curr_status}, for each of the regimes, we consider three standard benchmarks for VQA: VQAv2~\cite{goyal2017making}, TextVQA~\cite{singh2019towards} and \StvqaDatasetNoSpace~\cite{biten2019scene}, and two for CAP: COCO Captions~\cite{coco_cap} and TextCaps~\cite{sidorov2020textcaps}.
We report the performance on each benchmark and the non-weighted averaged one (\emph{combined evaluation}) to quantify the models' reasoning capabilities from both visual and textual information as a single number. For VQA, we calculate this score only on VQAv2 and TextVQA test sets.
Lastly, in \cref{sec:subset_section}, we present a new subset evaluation setting for scene-text VQA to measure the model's ability to answer questions requiring reasoning over all modalities simultaneously.
For all datasets, we extract OCR information using Amazon Text-in-Image\footnote{\url{https://docs.aws.amazon.com/rekognition/latest/dg/text-detection.html}}~\cite{scatter,textadain,semimtr,tts}. The supplementary materials list the implementation details, additional dataset information and for completeness, a comparison with other methods, disregarding the models' size.

\subsection{Visual Question Answering Experiments}
\label{sec:vqa_exp}
\begin{figure*}[t]
    \centering
    \includegraphics[width=0.99\textwidth]{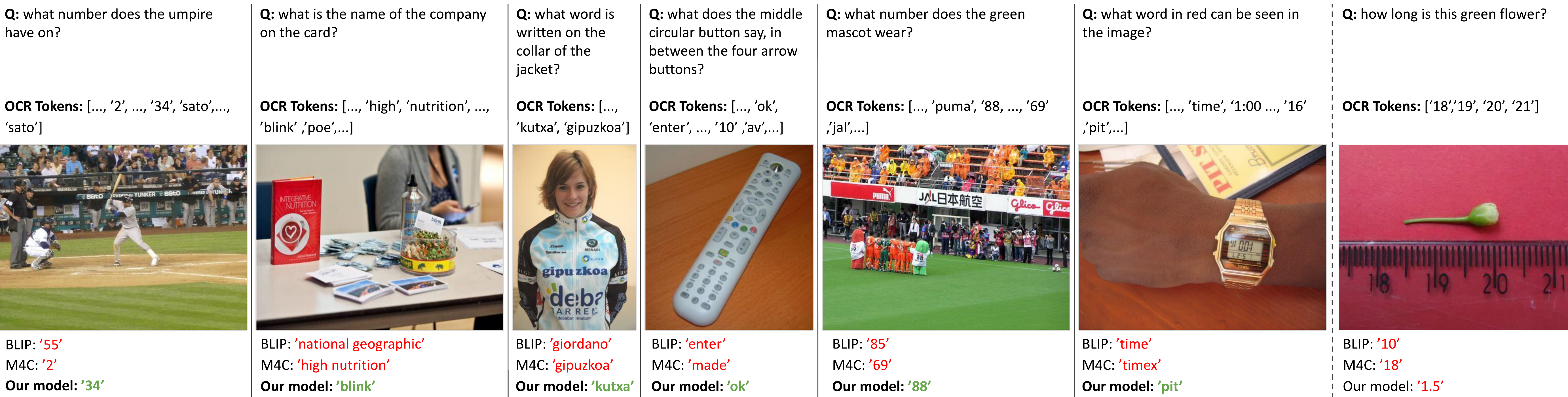}
    \caption{\textbf{Reasoning over all modalities.} We curate a subset out of TextVQA~\cite{singh2019towards}  validation set, containing only the samples which require reasoning over both vision and scene-text in the same question. Presented are representative examples from this subset, each includes an image, question, OCR input tokens, and model predictions. Green and red stand for correct and wrong predictions, respectively.}
    \label{fig:subset_exmaples}
    \vspace{-0.5cm}
\end{figure*}

We integrate our approach to two models, ALBEF and BLIP, denoted as \AlgoNameALBEF and \AlgoNameNoSpaceBLIP, respectively, and report their performance using three training regimes: (i) VQA, (ii) TextVQA, and (iii) \textit{Comb.}, as shown in \cref{table:vqa_textvqa}.
In the first regime, training \AlgoNameBLIP exclusively on VQAv2 results in performance improvements of $+3.19\%$ and $+12.16\%$ on VQA, and TextVQA, respectively, leading to a significant boost of $+7.67\%$ in the average score.
Even though VQAv2 mainly focuses on reasoning from visual information, these results stress the importance of scene-text understanding in this benchmark and the effectiveness of our method.
Interestingly, despite the marginal presence of OCR in VQAv2, \AlgoNameBLIP manages to effectively harness it and obtain $35.90\%$ on TextVQA, outperforming BLIP that trained solely on TextVQA itself ($27.72\%$).
In the scene-text configuration, performance improves by $+19.67\%$ on TextVQA; however, it decreases by $-3.15\%$ on VQA. 
This reinforces previous findings~\cite{wang2021towards,biten2022latr}, suggesting that \TextVqaTask datasets contain biases encouraging models to over-rely on the OCR and disregard the visual information. As BLIP's scene-text understanding is very restricted, it cannot fully exploit such biases and retains its visual understanding better, expressed via better VQAv2 results.
In the final combined training configuration, we showcase the versatility of our approach by presenting results for both \AlgoNameALBEF and \AlgoNameNoSpaceBLIP, highlighting its model-agnostic nature. 
When trained on both types of datasets, \AlgoNameBLIP improves BLIP by $+2.5\%$, $+22.77\%$, and $+13.28\%$ on VQA, TextVQA, and on average, respectively, achieving the highest average score.
The results indicate that despite the biases in \TextVqaTask datasets, \AlgoName can harness them without sacrificing the visual reasoning capability.
Moreover, \AlgoNameBLIP trained on the combined dataset outperforms models trained on each task separately, attesting to the tasks' mutually beneficial relationship, motivating the community to strive towards models that can see and read.

\begin{table}
\normalsize
\begin{center}

\small
\bgroup
\def\arraystretch{1.1}
    \resizebox{1\linewidth}{!}{%
        \begin{tabular}{l | cccc}
        \toprule
        \textbf{Method} & \textbf{TextVQA} &
        \textbf{$\text{TextVQA}_{\text{Read}}$} & \textbf{$\text{TextVQA}_{\text{See}\boldsymbol{\cap}\text{Read}}$}  & \textbf{Gap}$\downarrow$\\ 
        \hline
        M4C \cite{hu2020iterative} & 46.53 & 47.94 & 35.69 & 12.25\\
        TAP \cite{yang2021tap} & 54.71 & 56.24 & 35.83 & 20.41\\
        \hline
        \AlgoNameBLIP & \textbf{55.21} & \textbf{56.32} & \textbf{44.44} & \textbf{11.88} \\
        \bottomrule
    \end{tabular}
    }
\egroup
\tiny
\caption{\textbf{TextVQA splits.} Accuracy of leading scene-text VQA methods on the two non-overlapping subsets of TextVQA validation data, and the gap between them. 'See-$\boldsymbol{\cap}$-Read' refers to our subset, in which reasoning over all modalities  is needed for each sample. 'Read' stands for the rest of the TextVQA validation set.}
\label{tab:textvqa_subset}
\vspace{-0.75cm}
\end{center}
\end{table}

To gain a better understanding of the enhancements achieved by \AlgoName across both "see" and "read" datasets, we conducted a qualitative analysis of our method, comparing it to BLIP and M4C in \cref{fig:highlevel_examples_teaser}, \cref{fig:subset_exmaples} and in the supplementary materials. Our analysis indicates that the improvements observed in VQA are due to questions that necessitate reading and those that become easier to answer with OCR information. Regarding scene-text VQA, M4C struggles to reason from visual information, while \AlgoName excels in this regard, resulting in significant performance improvements on both TextVQA and \StvqaDatasetNoSpace.

\subsection{Image Captioning Experiments}

Similar to our VQA experiments, we evaluate the performance of \AlgoName on CAP by comparing it to top-performing methods using the same three training regimes (Caps, TextCaps, and \textit{Comb.}). We only integrate our approach to BLIP (\AlgoNameNoSpaceBLIP), as ALBEF was not applied to captioning.
As the results in \cref{table:textcap_cap} indicate, in the Caps regime, our approach slightly improves both the per-task and the average results.
Like in VQA, in the TextCaps training regime, \AlgoName results in significant gains over TextCaps ($+19.4$ CIDEr points) but a decline in COCO ($-14.4$ CIDEr points), compared to BLIP.
Moreover, while combined training leads to the best COCO results, the best TextCaps performance is achieved via designated TextCaps finetuning. This phenomenon aligns with the earlier findings by \cite{sidorov2020textcaps}, attributing it to the different nature of ground truth captions in the scene-text and general benchmarks (additional analysis appears in the supplementary materials).
Nevertheless, the combined trained \AlgoName leads to the best average score across all methods and regimes.

\begin{table}
\normalsize
\begin{center}

\small
\bgroup
\def\arraystretch{1.1}
    \resizebox{1\linewidth}{!}{%
        \setlength\tabcolsep{2pt}
        \begin{tabular}{cccccc | cc | cc}
        \toprule
        \multirow{2}{*}{\textbf{OCR-Sys}} & \multirow{2}{*}{\textbf{OCR-Enc}} & \multirow{2}{*}{\textbf{Fuse}} & \multirow{2}{*}{$\bm{\mathcal{L}_{OCR-LM}}$} & \multirow{2}{*}{$\bm{\mathcal{L}_{OCR-BC}}$} & \multirow{2}{*}{\textbf{2-D}} & \multicolumn{2}{c}{ \textbf{VQA}} & \multicolumn{2}{c}{ \textbf{CAP}} \\ 
        & & & & & & VQAv2 & TextVQA & COCO & TextCap \\
        \midrule
        \ding{55} & \ding{55} & \ding{55} &\ding{55}  &\ding{55} & \ding{55} & \normalsize77.40 & \normalsize32.43 & \normalsize133.4 & \normalsize101.4 \\
        \checkmark & \ding{55} & \ding{55} &\ding{55}  &\ding{55} & \ding{55} & \normalsize77.66 & \normalsize43.02 & \normalsize133.5 & \normalsize109.7 \\
        \checkmark & \checkmark & \ding{55}  & \ding{55}  & \ding{55} & \ding{55} & \normalsize78.41 & \normalsize46.13 & \normalsize133.5 & \normalsize110.4\\
        \checkmark & \checkmark & \checkmark  & \ding{55}  & \ding{55} & \ding{55} & \normalsize78.65 & \normalsize47.38 & \normalsize133.5 & \normalsize118.3 \\
        \checkmark & \checkmark & \checkmark & \checkmark  & \ding{55} & \ding{55} & \normalsize79.86 & \normalsize52.66 & \normalsize133.8 & \normalsize118.9 \\
        \checkmark & \checkmark & \checkmark & \checkmark  & \checkmark & \ding{55} & \normalsize79.81 & \normalsize52.66 & \normalsize- & \normalsize- \\
        \checkmark & \checkmark & \checkmark & \checkmark  & \ding{55} & \checkmark & \normalsize79.74 & \normalsize52.93 & \normalsize134.0 & \normalsize119.1 \\
        \checkmark & \checkmark & \checkmark & \checkmark & \checkmark & \checkmark & \normalsize79.90  & \normalsize55.21 & \normalsize- & \normalsize-\\
        \bottomrule
    \end{tabular}
    }
\egroup
\vspace{-0.2cm}
\tiny
\caption{\textbf{\AlgoName design choices.} 
\AlgoNameBLIP results on VQA and CAP w.r.t its different building blocks.
}
\label{tab:main_blocks_ablation}
\vspace{-0.6cm}
\end{center}
\end{table}

\subsection{A subset for Reasoning Over All Modalities}
\label{sec:subset_section}
As illustrated in \cref{fig:highlevel_examples_teaser}, VQA data is composed of three categories. Some questions can be answered using just vision ('see'), some by reasoning over the scene-text information only ('read'), and some require reasoning over both modalities at once ('see-$\boldsymbol{\cap}$-read').
Since most of the questions in current benchmarks fall either in the 'see' or 'read' category, unifying them is beneficial for testing methods' performance on the whole space, denoted by 'see-$\boldsymbol{\cup}$-read', eliminating the model's prior on whether a question is of type 'see' or 'read'. 
However, the more challenging and intriguing questions are the ones that require reasoning over scene-text and visual information altogether, denoted as 'see-$\boldsymbol{\cap}$-read'.
To provide a more reliable way to evaluate VQA models on this questions' category, we manually curate all such image-question pairs from the TextVQA~\cite{singh2019towards} validation set, producing an evaluation subset of 480 image-question pairs out of the total 5000 (±10\%).

This subset can serve as a foundation for measuring models' capabilities on what we believe are the more challenging questions that the research community should tackle.
In \cref{fig:subset_exmaples}, we depict examples from this subset alongside the prediction of M4C~\cite{hu2020iterative}, BLIP~\cite{li2022blip}, and \AlgoNameNoSpace.
This qualitative analysis confirms that both scene-text and general VQA models struggle to cope with this type of questions, while \AlgoName is substantially better.
Moreover, in \cref{tab:textvqa_subset} we report the quantitative results of leading scene-text-oriented methods and 
\AlgoName on the non-overlapping subsets of TextVQA validation set, \textit{i.e.,} the '$\text{TextVQA}_{\text{See}\boldsymbol{\cap}\text{Read}}$' subset and its complementary set, '$\text{TextVQA}_{\text{Read}}$', exposing the performance degradation that occurred on the former.
As these findings suggest, our method leads to the best performance, affirming that it is indeed better at reasoning on scene-text and visual information simultaneously.
Nevertheless, as can be seen, while \AlgoName is a step forward, there is still a big room to improve on these types of challenging questions.

\section{Ablation Studies}
\label{sec:ablations}
In this section, we study the effect of our key contributions and test the impact of freezing the vision encoder.

\noindent\textbf{Design Choices}:
We ablate \AlgoNameNoSpace's components on both the general and scene-text-oriented datasets in \cref{tab:main_blocks_ablation}, where all numbers are reported under the \textit{Comb.} settings.
Since the trends in CAP results are similar to VQA, we will focus on analyzing the latter.
First, we report the added performance of a naive approach -- simply inserting the OCR tokens as an additional input to BLIP's existing text encoder, similar to \cite{hu2020iterative,yang2021tap,biten2022latr}.
As seen in \cref{tab:main_blocks_ablation}, the accuracy on TextVQA improves by $+10.59\%$ (from $32.43\%$ to $43.02\%$) while improving VQA results by $+0.22\%$.
Our designated OCR encoder increases TextVQA performance to $46.13\%$ ($+3.11\%$) while obtaining an additional $+0.75\%$ gain in VQA.
Introducing our VL-OCR decoding scheme (denoted as ``Fuse'') boosts us to $47.37\%$ on TextVQA and an extra $+0.24\%$ on VQA.
Furthermore, using $\mathcal{L}_{\text{OCR-LM}}$ significantly improves TextVQA performance by $+5.22\%$ (from $47.38\%$ to $52.66\%$) while gaining an extra $+1.21\%$ on VQA.
Finally, the combination of $\mathcal{L}_{\text{OCR-BC}}$ with the 2-D information gets us to $55.21\%$ and $79.9\%$ on TextVQA and VQA.
Overall, \AlgoName leads to significant $+22.78\%$ and $+2.50\%$ improvements on TextVQA and VQA over the combined trained BLIP.
\noindent\textbf{The Effect of Freezing the Visual Encoder}:
\begin{table}[!t]
	\centering
	\resizebox{0.9\linewidth}{!}{%
	\begin{tabular}	{@{\extracolsep{1pt}}l c| c c c c | c}
        \toprule	 	
        \multirow{2}{*}{\textbf{Method}} & \multirow{2}{*}{\shortstack[c]{\textbf{Freeze} \\ \textbf{VE}}} &  \textbf{VQA} & \textbf{TextVQA} & \textbf{Avg.} \\
        & & test-dev & val \\
        \midrule
        \AlgoNameALBEF & \ding{55} & 75.55 & 40.60 & 58.08 \\ 
        \AlgoNameALBEF & \checkmark & 77.60 & 43.73 & 60.67 \\
        \cdashline{1-5}
        \AlgoNameBLIP & \ding{55} & 78.77 & 52.45 & 65.61 \\  
        \AlgoNameBLIP & \checkmark & 79.90 & 55.21 & 67.56 \\
	    \bottomrule
	\end{tabular}
	}
    \caption{\textbf{Visual encoder freezing.} VQA accuracy of \AlgoNameBLIP and \AlgoNameALBEF, with and without freezing the visual encoder, attesting to the freezing's importance.}
    \label{table:freezing_ablation}
    \vspace{-0.52cm}
\end{table}
Recently, a few works \cite{tsimpoukelli2021multimodal,alayrac2022flamingo,zhai2022lit} have examined different freezing configurations to avoid knowledge forgetness when combining pretrained models.
Inspired by these works, we examine the effect of freezing the Visual Encoder (VE) weights while applying \AlgoNameNoSpace, preserving its valuable knowledge acquired in pretraining, and summarize the results in \cref{table:freezing_ablation}.
As our findings suggest, freezing the VE significantly improves the results on VQA for both \AlgoNameALBEF and \AlgoNameBLIP by $+2.05\%$ and $+1.13\%$, and on TextVQA by $+3.13\%$ and $+2.75\%$, respectively.

\section{Discussion and Conclusions}
We wish to convey a few take-home messages to the VL research community. 
First, current SOTA methods cannot adequately reason over both scene-text and vision information.
Our experiments demonstrate that this occurs even when combining training datasets, suggesting a fundamental limitation of existing methods.
Second, our findings discover the symbiotic nature of these two types of reasoning capabilities, as performance on both tasks can be improved jointly. 
Moreover, by proposing \AlgoNameNoSpace, we present the first single model that successfully handles both task types.
Finally, we argue that the VL research community should strive to develop models that can simultaneously reason over vision, language, and scene-text.
To facilitate this, we curate a suitable subset to serve as a benchmark foundation. 




{\small
\bibliographystyle{ieee_fullname}
\bibliography{main_v2}
}

\clearpage

\appendix


\section{Implementation Details}
\label{app:implementation_details}
In this section, we provide full implementation specifics of UniTNT and divide it into three parts --  (1) architecture; (2) training procedure; (3) Scene-text information.

\subsection{Architecture}
We harness the model agnosticism of UniTNT and apply it to two top-performing VL models.
Specifically, we utilize the publicly-available code bases of ALBEF~\cite{albef}\footnote{\url{https://github.com/salesforce/ALBEF}} and BLIP\footnote{\url{https://github.com/salesforce/BLIP}}~\cite{li2022blip} and apply our method to them.
We design our approach in a modular way enabling simple integration into existing models.
Below we list the architectural specifics for both $\text{UniTNT}_{\text{ALBEF}}$ and $\text{UniTNT}_{\text{BLIP}}$.

\paragraph{OCR Encoder}
We use a pretrained BERT-base\footnote{\url{https://huggingface.co/docs/transformers/model_doc/bert}}~\cite{bert} as our encoder and introduce it with 2-dimensional information, as can be seen in Equation 1.
Specifically, we use three separate embedding layers (\textit{i.e.,} \texttt{torch.nn.Embedding})-- for the word token and its $x$ and $y$ axis positions for both the OCR and the question.
In particular, we define the minimal and the maximal spatial position as $0$ and $1000$, respectively, and set these values for the question tokens (referred to as ``pseudo-2D information'' in the main paper).
We restrict the number of OCR and question token lengths to $128$ and $35$, respectively.
Next, we sum the 2D-related embeddings and pass them in a 2-layer MLP with a hidden dimension of $768$ for additional processing.
Finally, we multiply it by $\alpha$ (set to $0.1$) and sum it with the token representation to obtain the final one fed into the encoder.

\paragraph{VL-OCR Decoder}
In order to introduce the pretrained decoder with scene-text information, we create new OCR Cross Attention (OCR-CA) blocks and place them in parallel to the existing VL ones.
Such newly added components are identical to the existing ones and initialized with the pretrained weights of the latters'.
To fuse the outputs of the OCR CA and the VL CA, $\mathcal{F}_{\text{OCR}}$ and $\mathcal{F}_{\text{VL}}$, we concatenate them along the channel dimension and pass them via attention based 2-layers MLP with a hidden size of $768$ to obtain $\mathcal{F}_\text{attn}$, an attention map that multiplies  $\mathcal{F}_{\text{OCR}}$ ($\mathcal{F}_{\text{OCR}}\odot \mathcal{F}_\text{attn}$).
Namely, this mechanism highlights the important and meaningful features in $\mathcal{F}_{\text{OCR}}$ and masks the less relevant ones.
Then, we pass the multiplication output via a learnable gating module (by multiplying it by $tanh(\beta)$, where $\beta$ is learnable and initialized to $0$), aimed to gradually blend the OCR features into the existing VL one.

\begin{figure}[!t]
\centering
    \includegraphics[width=0.5\textwidth]{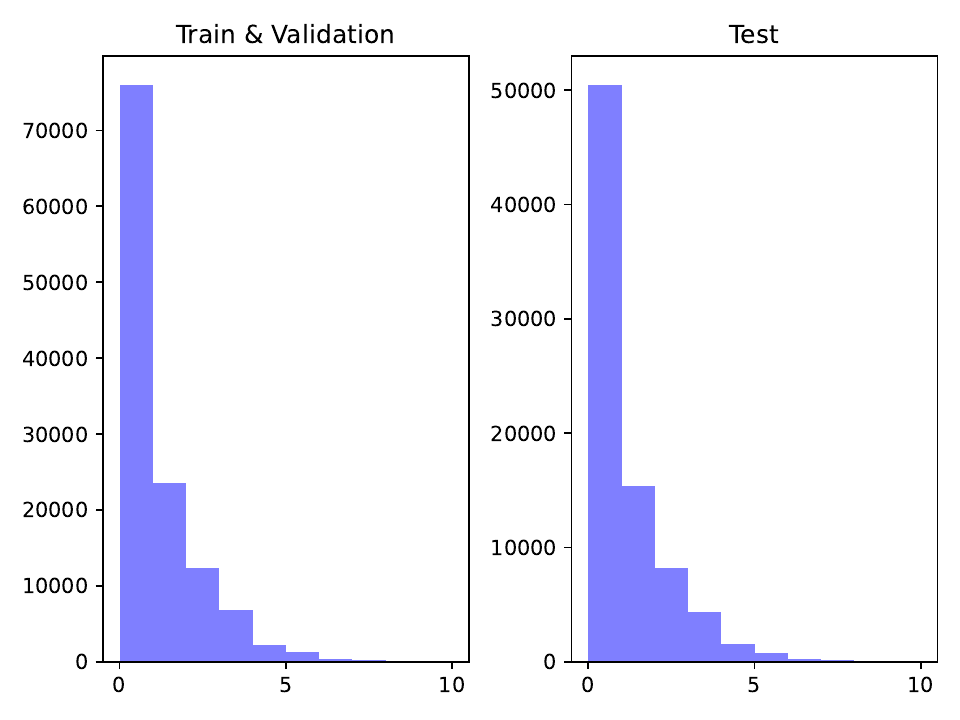}
    \caption{\textbf{OCR prevelance in VQAv2.} Histogram of the number of OCR instances per-image in VQAv2 dataset.}
    \label{fig:ocr_host}
\end{figure}

\subsection{Training Procedure}
We train all of our models to minimize $~{\mathcal{L}_{\text{\AlgoName}} = \mathcal{L}_{\text{base}} + \alpha_1\mathcal{L}_{\text{OCR-LM}} + \alpha_2\mathcal{L}_{\text{OCR-BC}}}$ using 8 A100 GPUs, where $\alpha_1$ and $\alpha_2$ are hyperparameters.

\paragraph{Visual Question Answering}
We train both $\text{UniTNT}_{\text{ALBEF}}$ and $\text{UniTNT}_{\text{BLIP}}$ on a unified Text-Non-Text VQA dataset, containing VQAv2~\cite{antol2015vqa}, TextVQA~\cite{singh2019towards} and ST-VQA~\cite{biten2019scene} for $10$ epochs using a batch size of $8$ and $16$ for ALBEF and BLIP, respectively.
Moreover, we set $\alpha_1=\alpha_2=1$ and keep the other training-related hyperparameters as in the original papers.

\paragraph{Image Captioning}
We train $\text{UniTNT}_{\text{BLIP}}$ on a the unified Text-Non-Text CAP dataset, comprised of COCO Captions~\cite{coco_cap} and TextCaps~\cite{sidorov2020textcaps}, for $5$ epochs with batch size of $32$. We set $\alpha_1 = 0.05$ and $\alpha_2 = 0$ since contrary to VQA, CAP does not contain textual information available both in training and inference time, making it infeasible to implement OCR-BC.
Moreover, we keep the rest of the hyperparameters as in BLIP.

\begin{table*}[t]
	\centering
	\resizebox{1\linewidth}{!}{%
	\begin{tabular}	{@{\extracolsep{1pt}} l c c | c c | c || c c | c}
        \toprule	 	
        \multirow{2}{*}{\textbf{Method}} & \multirow{2}{*}{\shortstack[c]{\textbf{Vision-oriented} \\ \textbf{dataset}}} & \multirow{2}{*}{\shortstack[c]{\textbf{OCR-oriented} \\ \textbf{dataset}}} &
        \multirow{2}{*}{\shortstack[c]{\textbf{VQA} \\ test-dev}} &
        \multirow{2}{*}{\shortstack[c]{\textbf{TextVQA} \\ val}} & \multirow{2}{*}{\textbf{Avg.}} & \multirow{2}{*}{\shortstack[c]{\textbf{COCO Caps} \\ val}} & \multirow{2}{*}{\shortstack[c]{\textbf{TextCaps} \\ val}} & \multirow{2}{*}{\textbf{Avg.}} \\
        & & & & & & &  \\ 
        \hline
        BLIP & \multirow{2}{*}{\ding{55}} & \multirow{2}{*}{\checkmark}  & 40.16 & 30.12 & 35.14 & 84.8 & 112.7 & 98.8 \\
        \AlgoNameBLIP  & & & 37.01 & 50.19 & 43.60 & 70.4 & \textbf{130.5} & 100.5 \\
        \cdashline{1-9}
        BLIP & \multirow{2}{*}{\checkmark} & \multirow{2}{*}{\ding{55}}  & 76.39 & 20.50 & 48.45 & 133.3 & 59.4 & 96.4\\
        \AlgoNameBLIP  & & & 79.68 & 36.33 & 58.01 & 133.7 & 59.6 & 96.7 \\
        \cdashline{1-9}
        BLIP & \multirow{2}{*}{\checkmark} & \multirow{2}{*}{\checkmark}  & 77.40 & 32.43 & 54.92 & 133.4 & 101.4 & 117.4 \\
        \AlgoNameBLIP  & & & \textbf{79.90} & \textbf{55.21} & \textbf{67.56} & \textbf{134.0} & 119.1 & \textbf{126.6}\\
        \bottomrule
	\end{tabular}
	}
	\caption{\textbf{The impact of training data.} We show the effect of each dataset configuration for training \AlgoName and BLIP.}
	\label{table:data_effects}
\end{table*}

\subsection{Scene-text information}
As specified in the paper, we extract the scene-text information (word tokens and 2-dimensional position) for all the VQA and CAP datasets (both the general and scene-text counterparts) using Amazon Text-in-Image.
To better understand the prevalence of OCR in the non-scene-text datasets, we plot the statistics of OCR in VQAv2 in \cref{fig:ocr_host} (same images are in COCO Captions as well).
While a large portion of the images does not contain text in them, there is a large amount of such with OCR ($38.36\%$ and $38.03\%$ of train and test images contain OCR).
Since OCR conveys meaningful information, it sheds light on the significant improvement of UniTNT up his baselines (ALBEF and BLIP).


\section{Datasets}
\label{app:datasets}

\subsection{Visual Question Answering}

\paragraph{VQAv2}
contains 204,721 images (82,783, 40,504, and 81,434) from COCO~\cite{lin2014microsoft}, 1,105,904 questions (443,757, 214,354, and 447,793), and 6,581,110 answers (4,437,570, 2,143,540, and the test answers are held-out).
Answering the questions requires vision-language understanding and commonsense knowledge.
Each question has ten ground-truth answers.

\paragraph{TextVQA}
contains 28,408 images from OpenImages~\cite{kuznetsova2020open}, 45,336 questions and 453,360 ground-truth answers.
The annotators were instructed to formulate questions that require reasoning from the text in the image.
As in VQAv2, each question has 10 ground-truth answers.

\paragraph{ST-VQA}
is a fusion of computer-vision datasets -- ImageNet~\cite{deng2009imagenet}, VizWiz~\cite{bigham2010vizwiz}, Visual Genome~\cite{krishna2017visual}, IIIT Scene Text Retrieval~\cite{MishraICCV13}, ICDAR 2013~\cite{karatzas2013icdar}, ICDAR 2015~\cite{karatzas2015icdar} and COCO Text~\cite{veit2016coco}.
It contains 31K questions, split into training (26K) and testing (5K), requiring scene-text understanding.

\subsection{Image Captioning} 

\paragraph{COCO Captions} contains over one and a half million captions describing over 330,000 images from the COCO dataset. 
Each image has five human-generated captions.

\paragraph{TextCaps}
is composed of 28,408 images and 142,040 captions (5 captions per image).
The images are from the TextVQA dataset, and the captions are based on the text in the image.
Specifically, models have to reason over the scene-text information to generate correct captions.

\section{The Impact of Training Data}
In this section, we study the effect of the different combinations of training datasets and report our findings in \cref{table:data_effects}.
In particular, we experiment with \AlgoName and BLIP in Visual Question Answering and Image Captioning using separate training on vision-oriented and OCR-oriented datasets and combined training.
In VQA, using both dataset types leads to the best standalone and average performance in the tested benchmarks.
This attests to the symbiosis between general and scene-text-oriented VQA, encouraging avoidance of the common practice of separate finetuning.

\begin{figure*}[!t]
\centering
    \includegraphics[width=\textwidth]{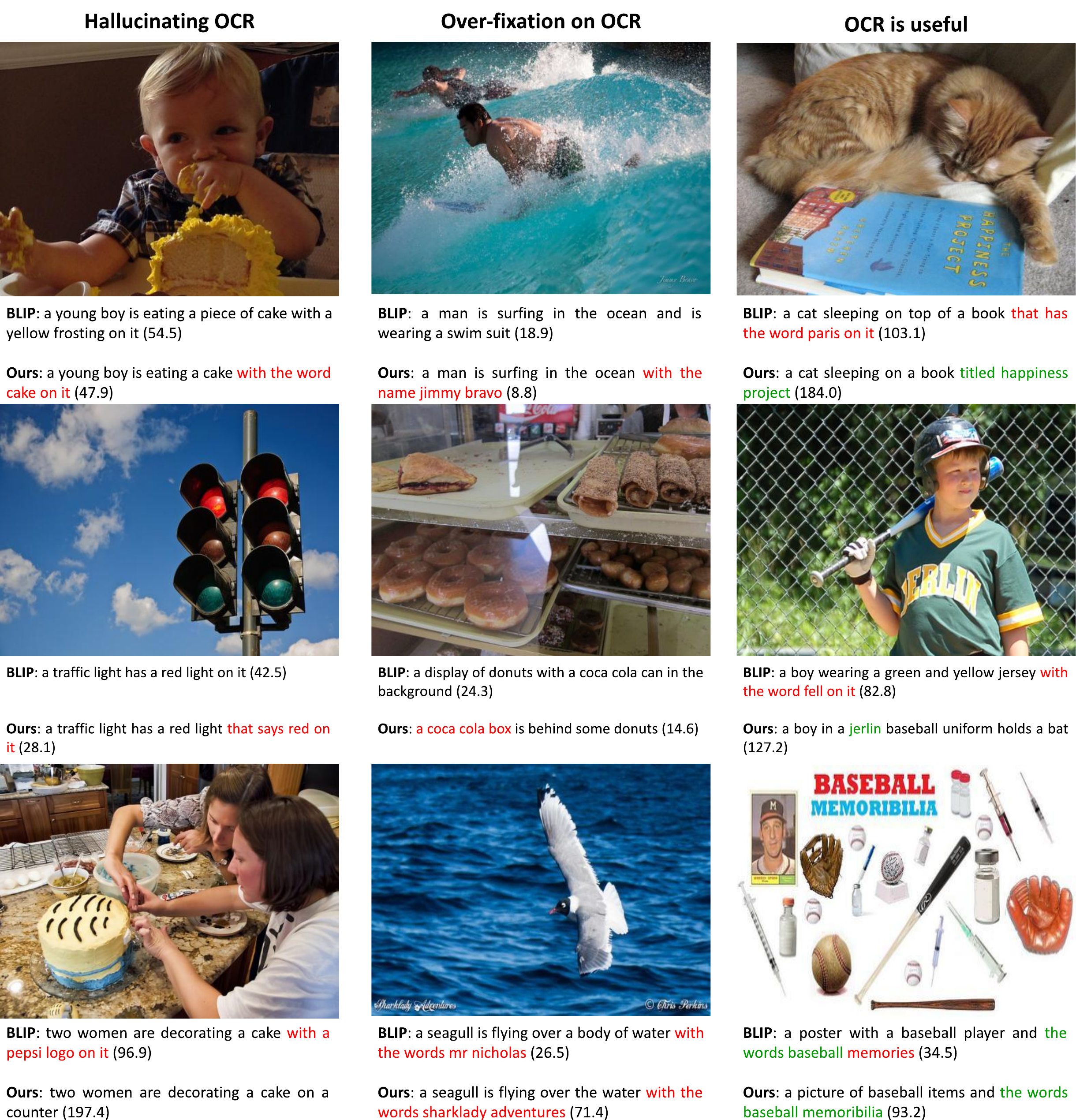}
    \caption{\textbf{Qualitative demonstration of the effects of finetuning on TextCaps.} BLIP and \AlgoName results of COCO Captions when finetuned solely on TextCaps. In some cases, scene-text understanding helps the models, but it also leads to over-reliance on the OCR signal and even to the hallucination of OCR. While such phenomena occur in both models, it is more prevalent in \AlgoName due to its better scene-text understanding.}
    \label{fig:textcaps_effect}
\end{figure*}

However, using a unified training set in CAP leads to the best COCO Captions and average results, but not in TextCaps.
Specifically, separate finetuning on TextCaps achieves a CIDEr score of $130.5$, compared to $119.1$ in the combined training.
It corresponds with \cite{sidorov2020textcaps}, which shows that combining COCO Captions with an upsampled version of TextCaps reduces the model's performance on the former.
It is because while training on TextCaps encourages the model to insert OCR into the caption, training on COCO Captions which barely contains OCR in its captions, penalizes such behavior, leading to an intrinsic tradeoff.
To better understand the effects of training models solely on TextCaps, we qualitatively test them on COCO Captions.
Notably, we finetune both BLIP and \AlgoName of TextCaps and demonstrate their performance on COCO Captions in \cref{fig:textcaps_effect}.
Our analysis shows that as TextCaps contains OCR in all its captions, separate finetuning causes models to fixate on OCR, regardless of their importance. Moreover, in images without an OCR signal, the models sometimes hallucinate text in the image.
While both models showcase similar behavior, since \AlgoName has better scene-text understanding, it is more prone to such phenomena.
It is also expressed in \cref{table:data_effects}, where BLIP and \AlgoName trained on TextCaps obtain $84.8$ and $70.4$ on COCO Captions, respectively.
Despite the improved performance on TextCaps when performing separate finetuning on it, our findings highlight its drawbacks. Thus, we claim that also in CAP, combined training should be applied.

From a general view, we hypothesize that since numerous valid captions exist for a given image, both with and without OCR, the model struggles to decide whether to use the OCR in its caption.
Due to the datasets' sizes in combined training that favors the vision-oriented ones, the model opts to reduce its use of OCR, not fully maximizing its performance on TextCaps.
It is contrary to VQA, where the conditioning over the question makes it easier for the model to decide whether to use OCR or not (\textit{e.g.}, \texttt{"What is written in the sign?"} versus \texttt{"What color is this shirt?"}).

\section{State-Of-The-Art Comparisons}
\label{app:sota_compare}

\begin{table*}[!t]
\centering
\begin{subtable}{.5\textwidth}
\centering
\captionsetup{width=.9\linewidth}
\begin{adjustbox}{max width=0.9\textwidth}
		\begin{tabular}	{@{\extracolsep{1pt}}l l c| c}
        \toprule
        & \multirow{2}{*}{\textbf{Method}} & \multirow{2}{*}{\textbf{\#Param.}} & \textbf{VQA}\\
        & & & test-std\\
        \midrule
        \parbox[t]{2mm}{\multirow{15}{*}{\rotatebox[origin=c]{90}{\textbf{\small Close}}}} 
        & OSCAR~\cite{li2020oscar} & - & 73.82 \\
        & UNITER~\cite{uniter} & - & 74.02 \\
        & VILLA~\cite{villa} & - & 74.87 \\
        & UNIMO~\cite{li2020unimo} & - & 75.27 \\
        & ALBEF~\cite{albef} & 300M & 76.04 \\
        & VinVL~\cite{zhang2021vinvl} & - & 76.60 \\
        & UFO~\cite{wang2021ufo} & - & 76.76 \\
        & CLIP-ViL~\cite{clipvil} & - & 76.70 \\
        & METER~\cite{meter} & - & 77.64 \\
        & BLIP~\cite{li2022blip} & 400M & 78.32 \\
        & $\text{SimVLM}_{\text{huge}}$~\cite{wang2021simvlm} & - & 80.34 \\
        & Florence~\cite{yuan2021florence} & 900M & 80.36 \\
        & $\text{mPlug}_{\text{VIT-L}}$~\cite{li2022mplug} & 600M & 81.26 \\
        & $\text{OFA}_{\text{huge}}$~\cite{ofa} & 900M & \underline{82.00} \\
        & CoCa~\cite{yu2022coca} & 2.1B & \textbf{82.3} \\
        \midrule
        \parbox[t]{2mm}{\multirow{4}{*}{\rotatebox[origin=c]{90}{\textbf{\small Open}}}} &
        Flamingo~\cite{alayrac2022flamingo} & 80B & \textbf{82.1} \\
        & GIT~\cite{wang2022git} & 700M & 78.81 \\
        \cdashline{2-4}
        &  $\text{UniTNT}_{\text{ALBEF}}$ & 400M & 77.80\\
        & \AlgoNameBLIP & 500M &  \underline{80.08}\\
	    \bottomrule
	\end{tabular}
\end{adjustbox}
\caption{Accuracy scores on VQAv2.}
\label{table:sota_comp_vqa}
\end{subtable}
\begin{subtable}{.5\textwidth}
\centering
\captionsetup{width=.9\linewidth}
\begin{adjustbox}{max width=0.9\textwidth}
		\begin{tabular}	{@{\extracolsep{1pt}}l l c | c c}
        \toprule	 	
        & \multirow{2}{*}{\textbf{Method}} & \multirow{2}{*}{\textbf{\#Param.}} &
        \textbf{TextVQA} & \textbf{ST-VQA}\\
        & & & test & test-ANLS\\
        \midrule
        \parbox[t]{2mm}{\multirow{5}{*}{\rotatebox[origin=c]{90}{\textbf{\small Close}}}} 
        & M4C~\cite{hu2020iterative} & 100M & 40.46 & 0.46 \\
        & LaAP-Net~\cite{han2020finding} & - & 41.41 & 0.49 \\
        & SA-M4C~\cite{kant2020spatially} & 200M & 44.60 & 0.50 \\
        & SMA~\cite{sma} & - & 45.51 & 0.47 \\
        & LOGOS~\cite{lu2021localize} & 100M & 51.08 & 0.58 \\
        & TAP~\cite{yang2021tap} & 150M & 53.97 & 0.60 \\
        \midrule
        \parbox[t]{2mm}{\multirow{4}{*}{\rotatebox[origin=c]{90}{\textbf{\small Open}}}} &
        LaTr-Large~\cite{biten2022latr} & 850M & \textbf{61.60} & \textbf{0.70} \\
        & Flamingo~\cite{alayrac2022flamingo} & 80B & 54.1 & - \\
        & GIT~\cite{wang2022git} & 700M & \underline{59.75} & \textbf{0.70} \\
        \cdashline{2-5}
        & $\text{UniTNT}_{\text{ALBEF}}$ & 400M & 44.13 & 0.58\\
        & $\text{UniTNT}_{\text{BLIP}}$ & 500M & 55.35 &  \underline{0.66} \\
	    \bottomrule
	\end{tabular}
\end{adjustbox}
\caption{Accuracy and ANLS scores on TextVQA and ST-VQA.}

\label{table:sota_comp_textvqa}
\end{subtable}
\caption{\textbf{VQA State-Of-The-Art comparison} The results of leading methods on the general and scene-text VQA benchmarks. While all other methods adopt a task-specific finetuning, UniTNT uses the same model for both, obtaining competitive results to much larger ones.}
\label{table:sota_comp1}
\end{table*}

\begin{table*}[!t]
\centering
\begin{subtable}{.5\textwidth}
\centering
\captionsetup{width=.9\linewidth}
\begin{adjustbox}{max width=0.9\textwidth}
		\begin{tabular}	{@{\extracolsep{1pt}}l c | c}
        \toprule	 	
        \multirow{2}{*}{\textbf{Method}} & \multirow{2}{*}{\textbf{\#Param.}} & \textbf{COCO}\\
        & & Karpathy-test\\
        \midrule
        MiniVLM~\cite{wang2020minivlm} & - & 119.8 \\
        DistillVLM~\cite{distillvlm} & - & 120.8 \\
        ViTCap~\cite{vitcap} & - & 125.2 \\
        OSCAR~\cite{li2020oscar} & - & 127.8 \\
        VinVL~\cite{zhang2021vinvl} & - & 130.8 \\
        UFO~\cite{wang2021ufo} & - & 131.2 \\
        BLIP~\cite{li2022blip} & 250M & 133.3 \\
        Flamingo~\cite{alayrac2022flamingo} & - & 138.1 \\
        LEMON~\cite{hu2022scaling} & - & 139.1 \\
        SimVLM~\cite{wang2021simvlm} & - & 143.3 \\
        CoCa~\cite{yu2022coca} & 2.1B & 143.6 \\
        $\text{OFA}_{\text{huge}}$~\cite{ofa} & 900M & \textbf{145.3} \\
        GIT~\cite{wang2022git} & 700M & \underline{144.8} \\
        \cdashline{1-3}
        \AlgoNameBLIP & 350M & 134.0\\
	    \bottomrule
	\end{tabular}
\end{adjustbox}
\caption{CIDEr scores on COCO Captions.}
\label{table:sota_comp_cap}
\end{subtable}
\begin{subtable}{.5\textwidth}
\centering
\captionsetup{width=.9\linewidth}
\begin{adjustbox}{max width=0.9\textwidth}
		\begin{tabular}	{@{\extracolsep{1pt}}l c | c}
        \toprule	 	
        \multirow{2}{*}{\textbf{Method}} & \multirow{2}{*}{\textbf{\#Param.}} & 
        \textbf{TextCaps}\\
        & & test\\
        \midrule
        BUTD~\cite{butd} & - & 33.8 \\
        AoANet~\cite{aoa} &  - & 34.6 \\
        M4C-Captioner~\cite{sidorov2020textcaps} & 100M & 81.0 \\
        Anc.-Captioner~\cite{anc} & - & 87.4 \\
        MMA-SR~\cite{wang2020multimodal} & - & 88.0 \\
        CNMT~\cite{anonymous2020challenge} & - & 93.0 \\
        TAP~\cite{yang2021tap} & 150M & 103.2 \\
        GIT~\cite{wang2022git} & 700M & \textbf{138.2} \\
        \cdashline{1-3}
        \AlgoNameBLIP & 350M & \underline{109.4} \\
	    \bottomrule
	\end{tabular}
\end{adjustbox}
\caption{CIDEr scores on TextCaps.}

\label{table:sota_comp_texcaps}
\end{subtable}
\caption{\textbf{CAP State-Of-The-Art comparison} The results of leading methods on the general and scene-text CAP benchmarks. While all other methods adopt a task-specific finetuning, UniTNT uses the same model for both, obtaining competitive results.}
\label{table:sota_comp2}
\end{table*}
In this section, we provide a complemental SOTA comparison to Tables 2 and 3, including additional models, regardless of their size, for presenting the complete performance status.
In these tables, we list the results reported in the respective papers alongside the number of trainable parameters.
While all other methods utilize different and distinct models for the general and the scene-text benchmarks (\textit{i.e.}, one for VQAv2 and one for TextVQA and ST-VQA), we are the first to present a unified model that performs well on both.
Moreover, while most methods adopt the closed-vocabulary evaluation, we adopt the open and more challenging one.
In \cref{table:sota_comp1}, we report the results for the general (\cref{table:sota_comp_vqa}) and the scene-text (\cref{table:sota_comp_textvqa}) VQA benchmarks.
Our findings attest that while UniTNT is the first to reason over both scene-text and visual information, it leads to comparable results with other, much larger models.

\section{Qualitative Analysis}
\label{app:qualitative_analysis}
\begin{figure*}[t]
    \centering
    \includegraphics[width=\textwidth,height=0.95\textheight,keepaspectratio]{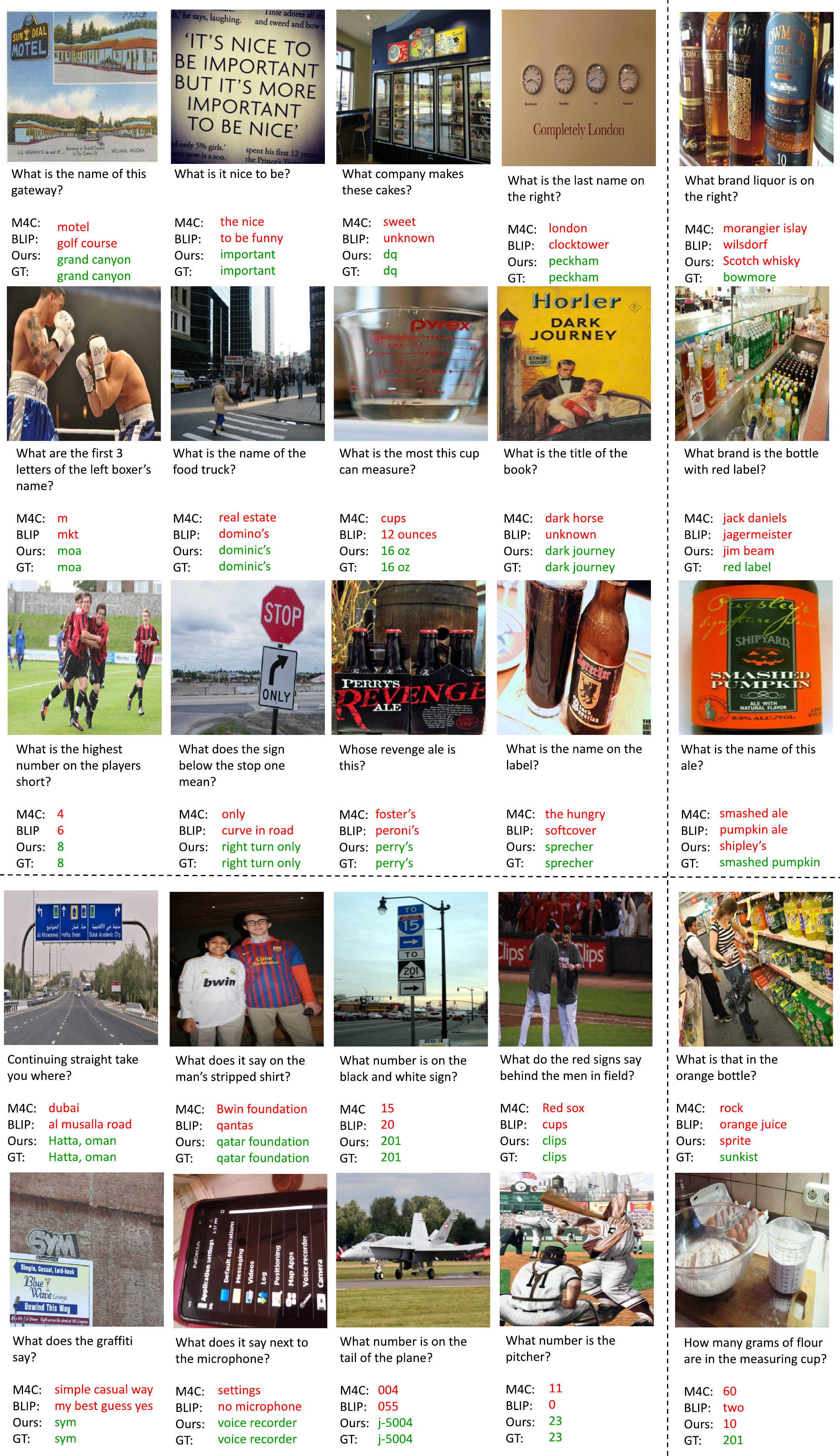}
    \caption{\textbf{Qualitative demonstration on TextVQA validation.} UniTNT, M4C, and BLIP answers, containing both success (left) and fail (right) cases of our method on image-question pairs that require mainly reading (top) and ones that require also visual reasoning (bottom).}
    \label{fig:qual_textvqa}
\end{figure*}

\begin{figure*}[t]
    \centering
    \includegraphics[width=\textwidth,height=0.95\textheight,keepaspectratio]{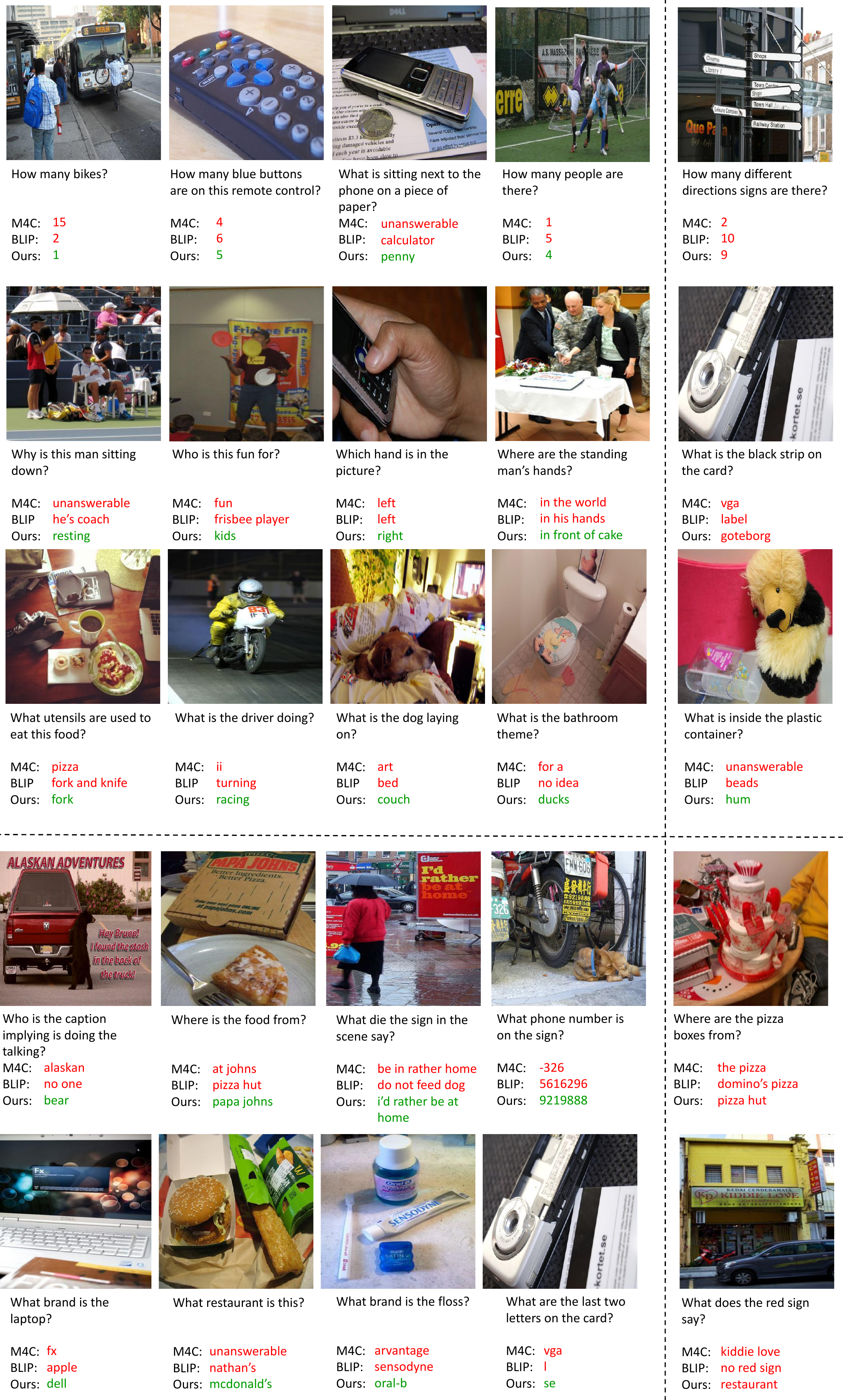}
    \caption{\textbf{Qualitative demonstration on VQAv2 test.} UniTNT, M4C, and BLIP answers, containing both success (left) and fail (right) cases of our method on image-question pairs that require mainly vision (top) and ones that require also scene-text understanding (bottom).}
    \label{fig:qual_vqa}
\end{figure*}

\paragraph{Visual Question Answering}
We provide an additional qualitative demonstration of UniTNT and compare it to BLIP and M4C on both TextVQA validation set (\cref{fig:qual_textvqa}) and VQAv2 test set (\cref{fig:qual_vqa}).
We depict in the four leftmost columns success-cases and the rightmost, fail cases, and color in green the correct answers and red, incorrect ones.
Moreover, we divide the figures such that the upper part corresponds with the benchmark's goal (VQAv2 -- see, TextVQA -- read) and the lower one with the counterpart goal (VQAv2 -- read, TextVQA -- see).
These results further demonstrate that UniTNT is capable of reasoning over both visual and scene-text information, while other competing methods perform unsatisfactorily on at least one of the benchmarks.
Moreover, as the visualizations in \cref{fig:qual_vqa} testify, granting scene-text understanding also benefit VQAv2, corresponding with the quantitative evidence in the main paper.
It is demonstrated in the bottom part of the figure, where the OCR is crucial for answering the questions or providing meaningful information that facilitates answering them.

\begin{figure*}[t]
    \centering
    \includegraphics[width=\textwidth,height=0.9\textheight,keepaspectratio]{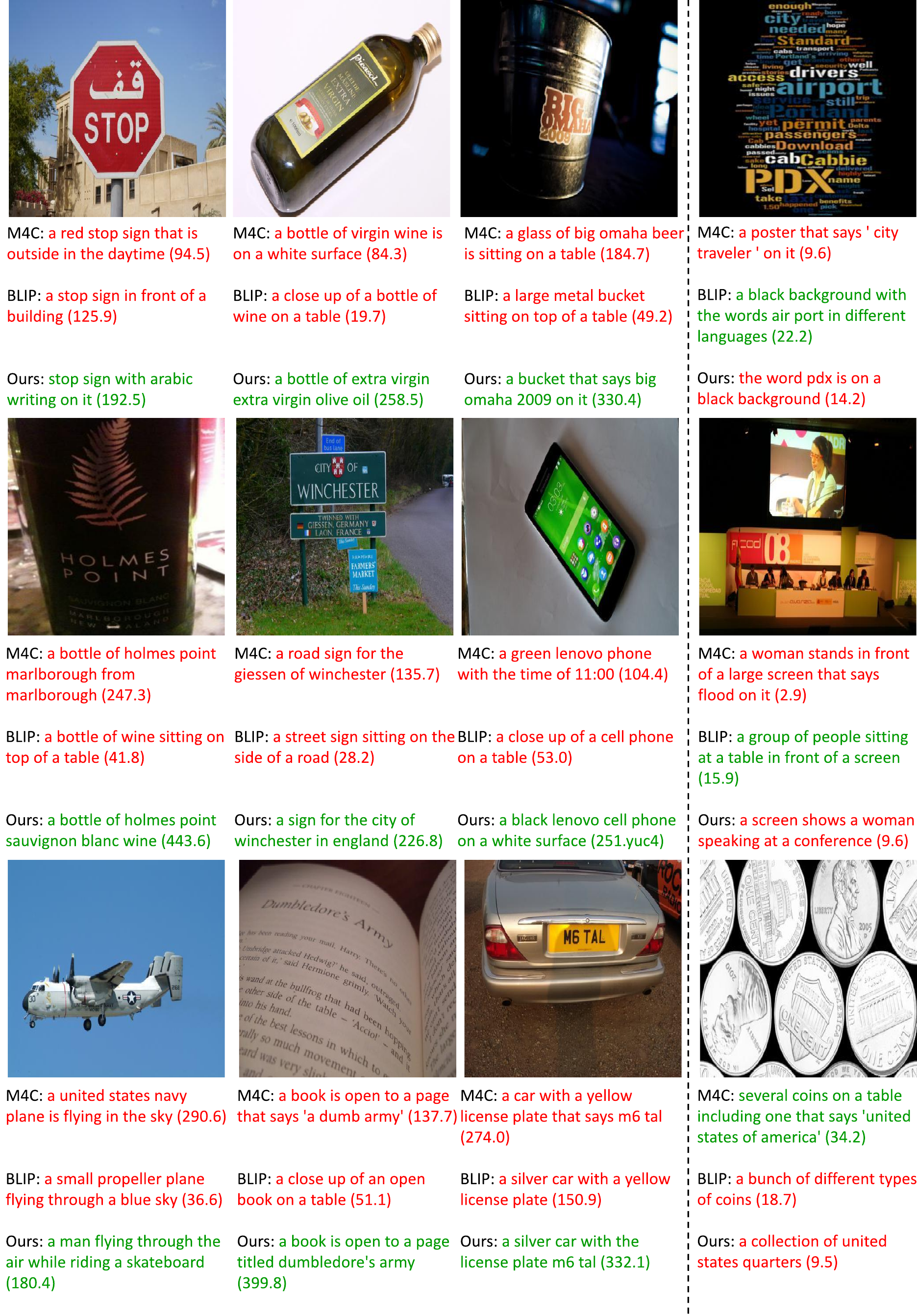}
    \caption{\textbf{Qualitative demonstration on TextCaps.} UniTNT, M4C-Captioner, and BLIP answers, containing both success (left) and fail (right) cases of our method alongside the per-caption CIDEr score.}
    \label{fig:qual_textcaps}
\end{figure*}

\begin{figure*}[t]
    \centering
    \includegraphics[width=\textwidth,height=0.9\textheight,keepaspectratio]{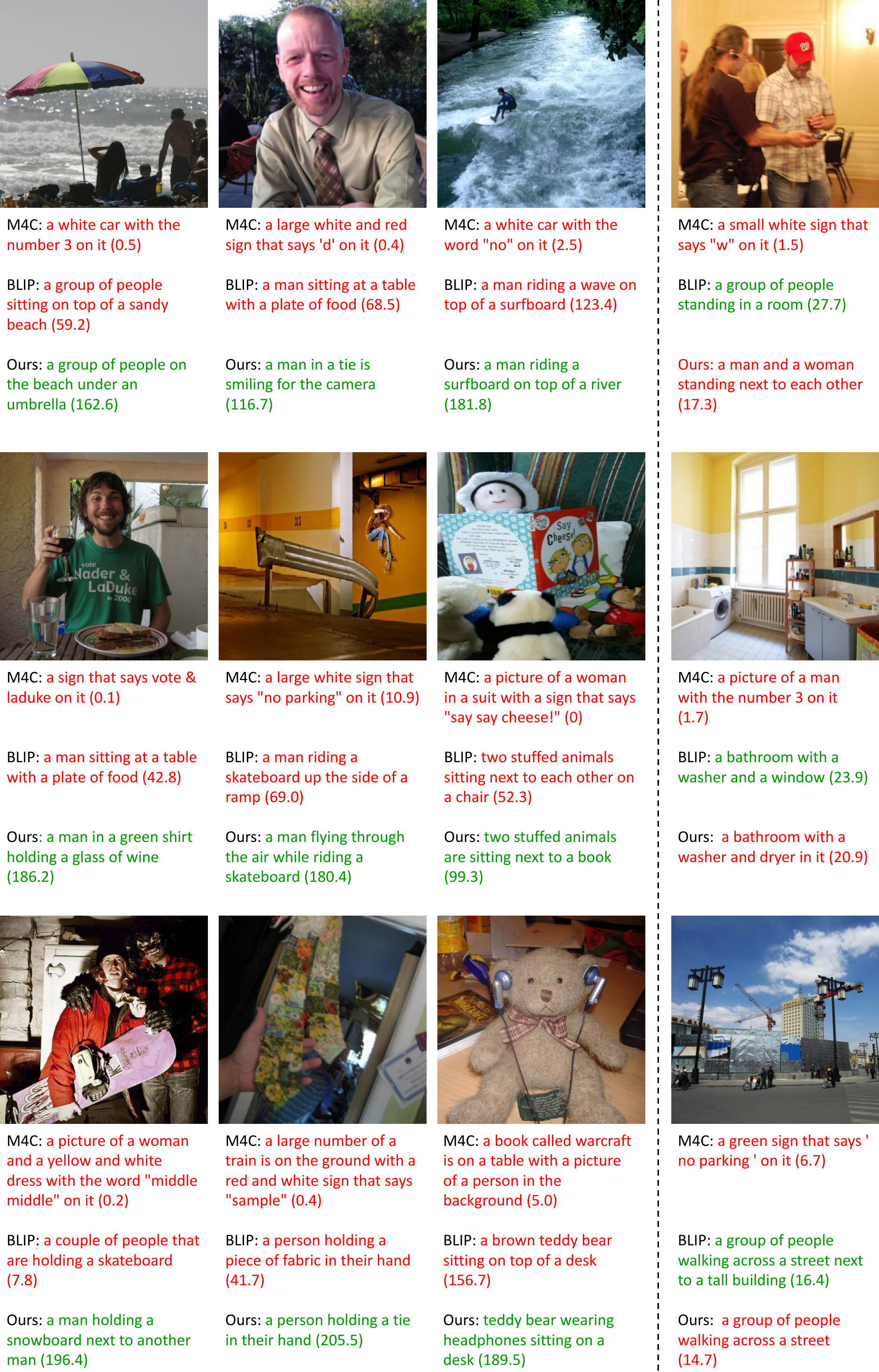}
    \caption{\textbf{Qualitative demonstration on COCO Captions.} UniTNT, M4C-Captioner, and BLIP answers, containing both success (left) and fail (right) cases of our method alongside the per-caption CIDEr score.}
    \label{fig:qual_cap}
\end{figure*}

\paragraph{Image Captioning}
Similar to the VQA demonstration, we present a visualization of UniTNT performance on TextCaps (\cref{fig:qual_textcaps} and COCO Captions (\cref{fig:qual_cap}) and compare the performance to M4C and BLIP.
On the left columns, we show images where our method outperforms the other methods, and on the right, its failure cases. 
Moreover, we list the CIDEr scores of each prediction and color in green the highest one.
These findings attest that BLIP is incapable of incorporating scene-text information, which results in relatively low CIDEr results.
Interestingly, M4C is too overfitted for TextCaps, causing it to fail completely on COCO Captions where OCR is scarce. Specifically, it focuses on the OCR regardless of their importance (\textit{e.g.}, the third example in the last row of \cref{fig:qual_cap}) and thus provides an irrelevant caption.
Despite the intrinsic tradeoff described in the paper between TextCaps and COCO Captions, UniTNT is capable of providing adequate captions for both benchmarks.
Specifically, our method is the only one to cope satisfactorily on both benchmarks altogether and is capable of harnessing both scene-text and visual information.

\end{document}